\documentclass[sensors,review,accept,moreauthors]{Definitions/mdpi}
\firstpage{1} 
\pubvolume{26}
\issuenum{20}
\articlenumber{6544}
\pubyear{2026}
\copyrightyear{2026}
\externaleditor{Bijan Shirinzadeh, Yongmin Zhong, Gaoge Hu and Guangle Gao}
\datereceived{16 August 2026} 
\daterevised{28 September 2026} 
\dateaccepted{10 October 2026} 
\datepublished{15 October 2026}
\soulregister\cite7
\renewcommand{\hl}[1]{#1}
\usepackage{comment}

\Title{Active Visual Perception: Opportunities and Challenges}

\Author{
\hl{Yian Li} $^{1}$\orcidA{}, 
Xiaoyu Guo $^{1}$,
Hao Zhang $^{1}$\orcidB{},
Jiayu Chen $^{1}$,
Shuiwang Li $^{1,}$*
and Xiaowei Dai $^{2,}$*
}

\AuthorNames{Yian Li, Xiaoyu Guo, Hao Zhang, Jiayu Chen, Shuiwang Li and Xiaowei Dai}
\address{%
$^{1}$ \quad College of Computer Science and Engineering, Guilin University of Technology, Guilin 541006, China; liyian@glut.edu.cn (Y.L.); 
guoxiaoyu@glut.edu.cn (X.G.); zhanghao@glut.edu.cn (H.Z.); \linebreak  
chennjiayu@163.com (J.C.)\\
$^{2}$ \quad College of Computer and Data Science, Putian University, Putian 351100, China
}
\corres{Correspondence: \hl{lishuiwang0721@163.com} 
 (S.L.); daixiaowei@ptu.edu.cn (X.D.)}

\abstract{Active visual perception enables an intelligent system to improve perception by controlling how visual evidence is acquired. However, differences in how studies define active perception and evaluate its benefits make approaches difficult to compare across tasks. This review first clarifies the field’s functional scope by examining whether an action is selected for its expected effect on subsequent visual evidence and whether the resulting observation informs a later decision. This definition distinguishes active observation acquisition from processing fixed inputs and guides the selection of representative literature. Within this scope, this review develops an analytical framework organized around four questions. It begins with what evidence the task requires, examines which actions can obtain that evidence and how a candidate observation should be valued, and considers how the next sensing action should be selected and executed. New observations then update the perceptual state and guide subsequent decisions, forming a closed loop that provides a common basis for comparing methods across applications. To examine how their benefits can be demonstrated, this review assesses whether existing datasets support observation selection, how interactive environments enable controlled comparisons, and what physical platforms reveal about execution feasibility. It further examines how baseline comparisons can establish whether adapting observations is necessary and whether informed selection offers benefits beyond simpler sensing strategies. These considerations underpin an evaluation framework that links improvements in visual evidence to downstream task outcomes while accounting for acquisition costs and risks. Its core value is to help isolate the contribution of sensing decisions and determine whether their benefits justify the resources required. The analysis concludes that predicted observation value does not guarantee that useful evidence can be acquired through feasible actions, while task success alone cannot establish the benefit of active observation selection. Further progress requires reliable value estimation and planning over longer decision sequences under physical and resource constraints. It also requires grounding semantic reasoning in acquired evidence and maintaining effective collaboration under imperfect communication, with these capabilities verified through reproducible evaluation of the complete perception and action loop.}


\begin{document}

\section{Introduction 
}
\label{sec:introduction}

Visual perception is commonly treated as a process that begins after images or videos have already been acquired~\cite{bajcsy2018revisiting}. Under~this assumption, the~main challenge is to extract reliable representations, estimates, or~predictions from a given input. This assumption is reasonable when the available observation contains sufficient evidence for the task, but~it becomes restrictive when the evidence is incomplete, ambiguous, or~poorly aligned with the current objective. A~manipulation system cannot reliably estimate a grasp region that remains fully occluded
~\cite{xiongvision,liu2026activevla}. An~autonomous vehicle cannot directly assess a road user that remains outside its observable field because of occlusion or limited sensing range~\cite{xu2023v2v4real}. A~tracking system cannot preserve target identity after visibility has been lost for a prolonged period, and~a reconstruction system cannot recover a surface that has not been observed from a useful direction
~\cite{murray1994motion,casao2024distributed,dunn2009next}. A~stronger perception model may improve the interpretation of available data, but~it cannot fully compensate for evidence that was never acquired. The~system must therefore decide not only how to interpret its current observation, but~also how to obtain a more useful~one.

Active visual perception introduces this decision into the perception process. An~active system controls some part of observation acquisition according to the current perceptual state or task requirement. The~controlled variable may be a camera pose, gaze direction, platform trajectory, optical setting, sensing modality, physical interaction, or~allocation of sensing responsibility among several agents
~\cite{bajcsy2018revisiting,chen2011active}. The~selected action changes the evidence available to the system, and~the resulting observation influences later estimation, planning, communication, or~task execution. Perception guides the sensing action, while the acquired observation updates the state on which later actions depend. Bajcsy described active perception through the purposeful selection and control of data acquisition strategies~\cite{bajcsy1988active}. Aloimonos~et~al. further emphasized that visual perception can be improved by actively controlling the sensing process and the geometry of observation~\cite{aloimonos1988active}. Ballard similarly emphasized controllable gaze and camera coordinates as essential components of visually guided behavior~\cite{ballard1991animate}. Subsequent work established that selective observation can reduce ambiguity and avoid unnecessary processing when different measurements do not provide equal value to the current task~\cite{swain1993promising,bajcsy2018revisiting}.

This principle now appears across a wide range of intelligent systems. Active reconstruction selects viewpoints that expose unobserved surfaces and improve geometric completeness~\cite{dunn2009next,peralta2020next,chen2024gennbv}. Active localization, mapping, and~exploration use camera or platform motion to reduce localization and mapping uncertainty, improve spatial coverage, and~acquire more informative observations
~\cite{leung2006active,placed2022explorb,bonetto2022irotate}. Visual search and active recognition acquire additional observations when the location, identity, or~relevance of a target remains uncertain~\cite{laporte2006efficient,taioli2024unsupervised,bajpai2025uncertainty}. Active tracking redirects a camera, transfers responsibility between cameras, or~changes platform motion to preserve visibility under target motion and occlusion~\cite{murray1994motion,rivlin2000control,casao2024distributed}. Robotic manipulation may move the camera, reposition the robot, or~interact physically with an object when the current view does not reveal the grasp region, contact state, or~task progress~\cite{le2008active,krainin2011autonomous,weng2024interactive}. Collaborative systems may coordinate which agent acquires complementary evidence or selectively communicate perceptually critical information when the local observation is insufficient
~\cite{queralta2020collaborative,hu2022where2comm}.

Although these systems pursue different task outcomes, they repeatedly face the same technical dependency. A~sensing action can be useful only when the system has identified what the current observation fails to provide. The~system must then determine which controllable action can expose the required evidence, predict whether the resulting observation will improve perception or task execution, and~decide whether the expected improvement justifies the associated cost. These decisions are often described through different task-specific terms. Reconstruction emphasizes coverage and completeness, localization emphasizes uncertainty and observability, search emphasizes target discovery, tracking emphasizes visibility preservation, manipulation emphasizes task success, and~collaborative perception emphasizes complementary evidence and communication efficiency~\cite{dunn2009next,placed2022explorb,taioli2024unsupervised,casao2024distributed,krainin2011autonomous,hu2022where2comm}. Similar mathematical quantities can therefore represent different forms of observation value. Information-based objectives, for~example, can refer to uncertainty reduction in mapping or recognition, while task-oriented observation value may instead concern grasp verification or the usefulness of communicated evidence~\cite{laporte2006efficient,mu2016information,hu2022where2comm}.

{\hl{Existing} 
 reviews have established important foundations in active perception, sensor planning, robotic active vision, view planning, and~active SLAM~\cite{tarabanis1995survey,scott2003view,bajcsy2018revisiting,chen2011active,zeng2020view,ahmed2023active,placed2023survey,maboudi2023review}. Despite their different scopes, these reviews share several limitations from the perspective of cross-task active visual perception. First, most of them organize the literature around a specific application domain or sensing problem, such as robotic active vision, view planning, or~active SLAM. Consequently, methods that address similar observation--acquisition problems through different sensing actions are rarely compared within a common framework. Second, existing taxonomies mainly emphasize conceptual principles, sensor configurations, viewpoint-planning algorithms, or~task-specific uncertainty criteria, but~do not consistently connect the perceptual evidence missing from the current state, the~action used to acquire that evidence, the~expected value of the resulting observation, and~the subsequent state update as a unified closed loop. Third, existing evaluations generally focus on final task performance or planning efficiency, with~limited distinction among improvements in perceptual evidence, gains in downstream task performance, and~the feasibility and resource costs of acquiring new observations. These shared limitations make it difficult to determine whether a reported improvement originates from a better observation--acquisition decision or from other components of the perception and decision-making pipeline.}

To address these gaps, this review takes the observation--acquisition process, rather than a particular application domain or algorithm family, as~the unit of cross-task comparison. Representative methods are examined according to four linked elements, including the perceptual evidence missing from the current state, the~controllable action used to change the observation, the~expected value of the resulting evidence, and~the mechanism used to select, execute, and~incorporate the action into the next system state. This structure is further connected to evaluation by distinguishing improvements in the acquired evidence from improvements in downstream task performance and from the feasibility and resource costs required to obtain those improvements. Table~\ref{tab:prior_surveys} summarizes how this perspective complements representative previous~reviews.

The practical consequence is that reported performance gains can be difficult to attribute to the sensing decision itself. For~example, a~predicted camera pose may appear informative while remaining physically unreachable for the platform~\cite{zeng2020view,jauhri2024active}. Likewise, improved final task performance does not necessarily demonstrate that a learning policy has acquired more useful visual evidence. In~collaborative perception, higher detection accuracy may result from receiving additional information without showing that the message selection strategy itself is more effective than uniform communication~\cite{hu2022where2comm,yang2023how2comm,huang2024actformer}. These examples illustrate several possible failure points in the observation acquisition process. A~system may misidentify the required evidence, select an unsuitable sensing action, misestimate the value of a candidate observation, or~fail to execute the selected action under physical and resource constraints. A~review centered on these dependencies is therefore needed to compare methods across tasks and to determine whether their reported benefits arise from effective active observation~selection.

\begin{figure}[H]
\includegraphics[width=\linewidth]{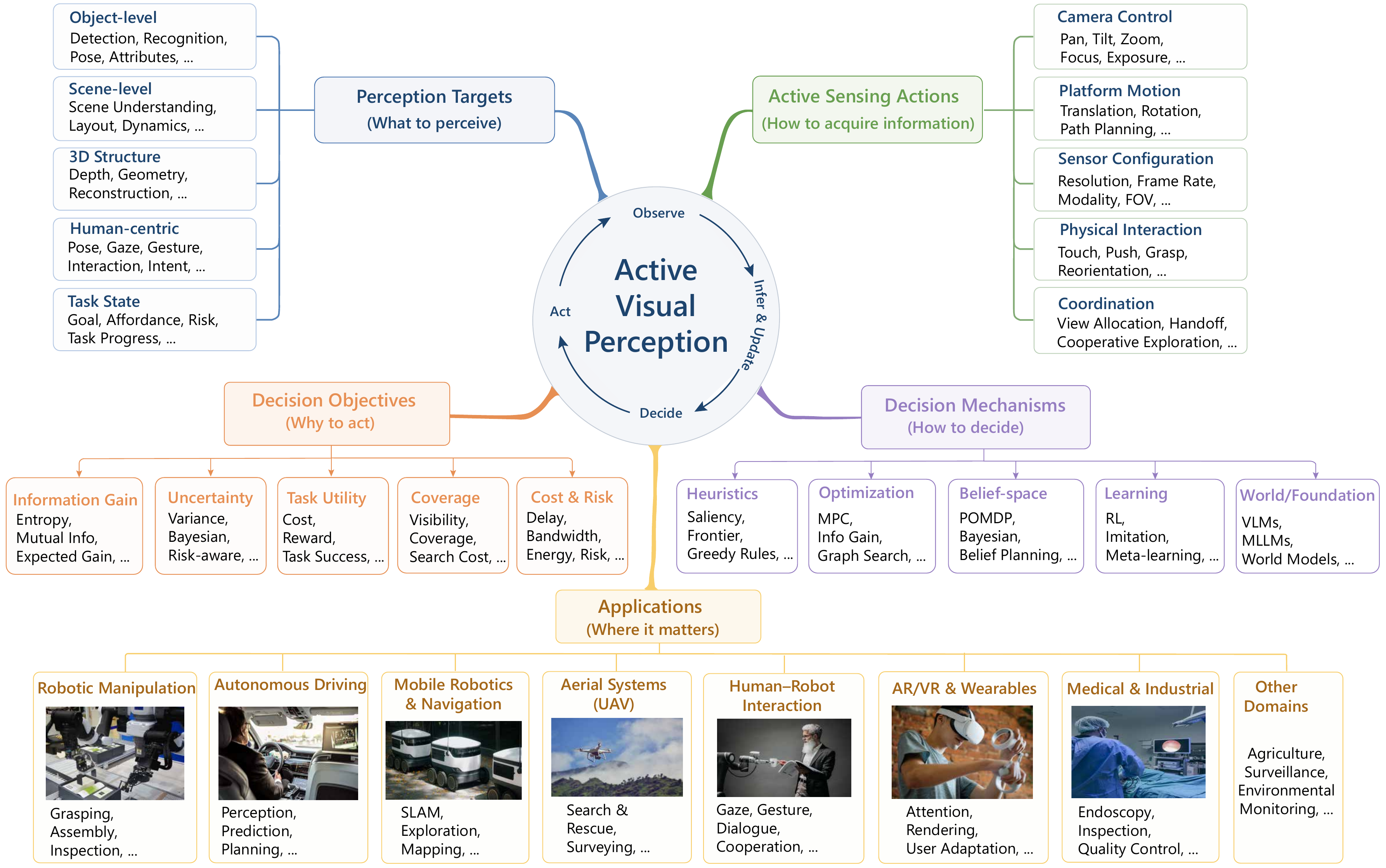}
\caption{\hl{Conceptual} 
 framework of active visual perception linking perception targets, sensing actions, decision objectives, decision mechanisms, and~application~domains.}
\label{fig:framework}
\end{figure}

Such distinctions become even more important as active perception is increasingly integrated with vision language and vision language action models. These models can connect instructions with objects, regions, spatial relations, and~task states, and~they can use semantic reasoning to guide navigation, manipulation, or~camera motion toward potentially relevant evidence~\cite{bajpai2025uncertainty,xue2026profocus,zhang2026embodied,liu2026activevla,liu2026sapave}. Their semantic capability does not remove the classical active perception problem. A~semantically plausible viewpoint may still be physically unreachable, unsafe, visually uninformative, or~too expensive to obtain. A~model may also produce a confident interpretation that is not adequately supported by the current observation. Active perception with foundation models must therefore connect semantic relevance with observability, uncertainty, feasibility, and~evidence verification. Figure~\ref{fig:framework} provides an overview of the main perception targets, sensing actions, decision objectives, decision mechanisms, and~application domains considered in active visual~perception.

\begin{figure}[H]
\includegraphics[width=\linewidth]{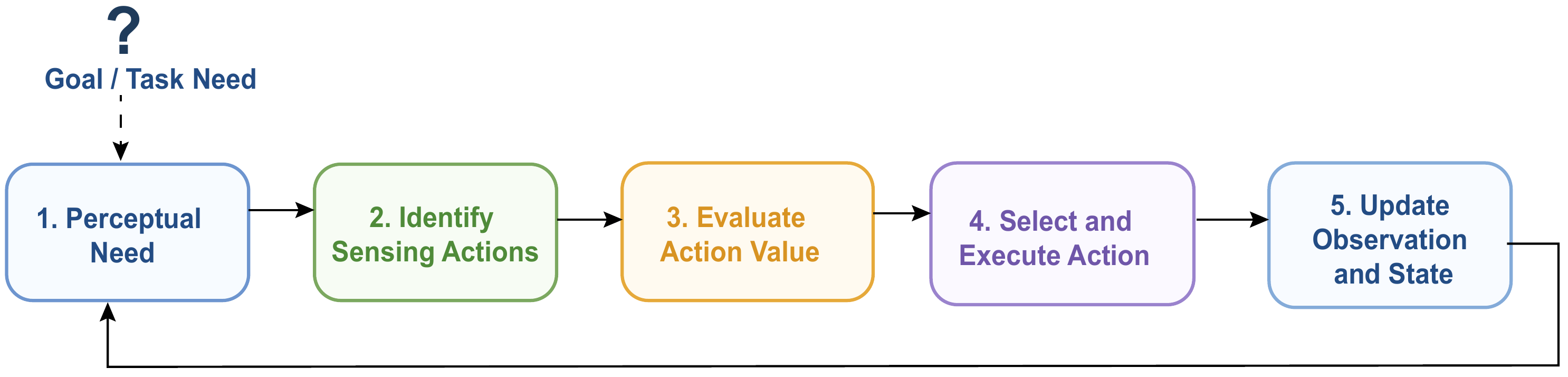}
\caption{{Closed-loop} 
 process linking perceptual needs, sensing actions, action evaluation, execution, and~state~updates.}
\label{loop}
\end{figure}

This review examines active visual perception through four connected questions. First, the~system must determine what visual evidence is required by the current task. Second, it must determine which sensing or embodied actions can change the observation and expose that evidence. Third, it must estimate the value of the candidate observation by considering its expected effect on perception and task execution together with its acquisition cost. Fourth, it must select and execute the next sensing action. After~the new observation becomes available, the~system updates its perceptual and task state, which closes the loop and determines the remaining perceptual need. These four questions describe successive decision dependencies, while the state update provides the feedback required for continued sensing. They do not assume that methods belong to mutually exclusive algorithmic categories. Geometric reasoning, probabilistic estimation, information-based planning, reinforcement learning, foundation models, and~distributed coordination can each contribute to one or more stages of the loop. Figure~\ref{loop} summarizes these dependencies. The~selected sensing action changes the next observation, while the acquired evidence updates both the perceptual state and the remaining task~requirement.

Based on the above definition and literature coverage, the~main contributions of this review are summarized as follows:

\begin{itemize}
    \item It clarifies the functional boundary of active visual perception by distinguishing it from passive visual processing, predetermined movement, fixed multimodal fusion, and~general active learning. The~defining criterion is whether an action is selected partly for its expected effect on subsequently acquired visual~evidence.

    \item It presents active visual perception as a closed-loop process involving four connected questions. These questions concern what visual evidence is required by the task, which actions can change the current observation, how the value of a candidate observation should be estimated, and~how the next sensing action should be selected and executed. This perspective supports comparison across reconstruction, localization, search, tracking, manipulation, sensor selection, and~collaborative~perception.

    \item It examines representative datasets, interactive environments, physical platforms, baselines, and~evaluation criteria. The~resulting evaluation perspective distinguishes improvement in the acquired visual evidence from improvement in the downstream task and from the motion, computation, communication, energy, time, and~safety costs required to obtain that~improvement.

    \item It identifies recurring challenges and research opportunities across active visual perception systems. These issues include task-guided evidence acquisition, uncertainty calibration, long-horizon decision-making, semantic verification, physical execution, collaborative sensing under imperfect communication, safety, and~reproducible closed-loop evaluation.
\end{itemize}

\begin{table}[H]
\caption{\hl{Comparison} 
 between representative previous reviews and the analytical scope of this~review.}
\label{tab:prior_surveys}

\small
\setlength{\tabcolsep}{3pt}
\renewcommand{\arraystretch}{1.25}

\begin{tabular}{
    >{\raggedright\arraybackslash}m{0.125\textwidth}
    >{\raggedright\arraybackslash}m{0.22\textwidth}
    >{\raggedright\arraybackslash}m{0.19\textwidth}
    >{\raggedright\arraybackslash}m{0.19\textwidth}
    >{\raggedright\arraybackslash}m{0.20\textwidth}
}
\toprule

\textbf{Review} &
\textbf{Primary Scope} &
\textbf{Observation Control} &
\textbf{Decision Emphasis} &
\textbf{Evaluation / Constraints} \\

\midrule

Bajcsy~et~al.~\linebreak  \cite{bajcsy2018revisiting} &
Conceptual foundations of active perception &
What, where, when, and~how to sense &
Task and context dependent sensing &
Resource constraints discussed conceptually \\
\midrule
Chen~et~al.~\linebreak  \cite{chen2011active} &
Active vision in robotic systems &
Sensor pose, configuration, and~viewpoint planning &
Uncertainty and task-oriented sensor planning &
Robot configurations, applications, and~method evaluation \\
\midrule
Zeng~et~al.~\linebreak  \cite{zeng2020view} &
View planning for robotic active vision &
Primarily sensor viewpoint and view sequence &
View quality and planning strategies &
Hardware conditions, scanning state, and~planning efficiency \\
\midrule
Ahmed~et~al.~\linebreak  \cite{ahmed2023active} &
Active and collaborative SLAM &
Robot trajectory and control action selection &
Pose and map uncertainty, information utility &
Planning methods, deployment settings, and~SLAM performance \\
\midrule
\textbf{\hl{This review} 
} &
\textbf{\hl{Cross-task active observation acquisition}} &
\textbf{\hl{Viewpoint, gaze, sensing, interaction, and~coordination}} &
\textbf{\hl{Required evidence, observation value, and~action selection}} &
\textbf{\hl{Evidence improvement, task improvement, feasibility, and~resource cost}} \\

\bottomrule
\end{tabular}
\end{table}

\vspace{-12pt}

The remainder of this review is organized as follows: Section~\ref{sec:definitions_scope} defines active visual perception, specifies its functional boundary, and~explains the study selection criteria. Section~\ref{sec:taxonomy} examines the perceptual evidence required by the task, the~actions that change visual observation, the~value of the acquired observation, and~the selection of the next sensing action. Section~\ref{sec:evaluation} discusses datasets, interactive environments, physical platforms, baselines, metrics, and~closed-loop evaluation requirements. Section~\ref{sec:applications} examines how the four parts of the sensing process appear in representative application settings. Section~\ref{sec:challenges} analyzes recurring limitations across existing systems. Section~\ref{sec:future_directions} develops research directions from these limitations, and~Section~\ref{sec:conclusion} concludes this~review.

\section{Definitions and~Scope}
\label{sec:definitions_scope}
\unskip

\subsection{Definition of Active Visual~Perception}

Not every change in visual input constitutes active visual perception. A~mobile robot, for~example, continuously receives new images while following a route, but~these images may simply be a by-product of movement. The~sensing process becomes active only when the robot chooses its action partly because of the visual evidence that the action is expected to provide. This distinction follows the early view of active perception as the purposeful control of sensing. Bajcsy emphasized that perception should include the selection and control of data acquisition strategies, while Ballard-related visual perception to controllable gaze and visually guided behavior~\cite{bajcsy1988active,ballard1991animate}. Later work further described an active agent as a system that determines what information should be acquired, as~well as where, when, and~how it should be obtained~\cite{bajcsy2018revisiting}.

Following this principle, this review defines active visual perception as a task-guided process in which a system selects an action partly for its expected effect on subsequent visual evidence. The~system may be a robot, a~vehicle, a~camera network, or~a group of sensing agents. What matters is that visual information is relevant to the current task, that the action responds to the current perceptual or task state, and~that the resulting observation influences a later estimate, plan, or~action. A~predetermined camera trajectory therefore does not become active merely because it produces a sequence of different views. The~same trajectory becomes relevant to active visual perception when it is chosen to reveal an occluded region, improve localization, preserve target visibility, or~acquire evidence required for task completion. The~defining property lies in the reason for selecting the action and in the role of the observation obtained after that~action.

Within this visually grounded formulation, active perception is not restricted to a single imaging modality. Visual evidence may come from RGB, depth, thermal, event, or~other imaging sensors, and~other measurements may support the visual decision. Non visual modalities are included in this review only when they complement, guide, or~are jointly selected with visual observations for a perceptual or task decision in which visual information remains an essential component. Systems that perform active sensing entirely through non visual measurements without affecting visual evidence acquisition are outside the present scope. A~system may combine several modalities because different sensors remain reliable under different environmental conditions. However, using several sensors does not by itself make perception active. The~process becomes active when the system decides which sensing modality should complement the visual observation, which region should be sampled, which viewpoint should be acquired, or~which information should be transmitted. A~system that always combines the same predetermined measurements performs multimodal fusion, whereas a system that changes its sensing resources according to the current information need performs active observation selection. The~same distinction applies to recent vision language systems. Language and semantic reasoning become relevant when they guide the acquisition of additional evidence, such as deciding where the system should look, move, revisit, or~interact. A~model that reasons only over a fixed image changes the interpretation of existing evidence, but~it does not change how new evidence is~obtained.

\subsection{Included Forms of Observation Control and~Tasks}

The definition above leads to several ways in which a system can improve its observations. The~most direct approach is to change the geometric relationship between the sensor and the scene. Camera translation, rotation, gaze control, viewpoint selection, zoom, and~viewing distance determine which regions are visible, how large an object appears, and~which geometric constraints can be recovered from the image. This form of control is central when the missing evidence can be obtained from another direction or scale. When changing the viewpoint alone is insufficient, the~system may instead change the sensing configuration. Exposure, focus, resolution, frame rate, field of view, sampling density, and~modality selection affect the quality, frequency, and~type of evidence that can be acquired. These decisions are especially important when observation quality varies with illumination, motion, distance, or~sensor~limitations.

Some information cannot be obtained by changing the camera settings alone because the sensor must move through the environment or the scene itself must be changed. Mobile robots may therefore select trajectories that improve feature visibility, reduce localization uncertainty, extend spatial coverage, or~maintain a target within view. Manipulation systems may go further by pushing, rotating, grasping, or~repositioning an object so that hidden surfaces or task-relevant regions become observable~\cite{le2008active,krainin2011autonomous,weng2024interactive}. In~distributed systems, the~missing evidence may already be visible to another sensor or agent. The~decision then concerns which agent should observe a region, which camera should continue tracking a target, which information should be transmitted, and~when sensing responsibility should be transferred~\cite{casao2024distributed,queralta2020collaborative,hu2022where2comm}. Camera control, platform motion, physical interaction, and~collaborative sensing therefore represent different ways of addressing the same underlying problem, which is how to change the available evidence when the current observation is~insufficient.

These forms of observation control recur across several research tasks. In~active reconstruction, the~system selects new views because the current model contains unobserved or poorly reconstructed surfaces~\cite{dunn2009next,peralta2020next,chen2024gennbv}. In~active localization and mapping, movement is selected not only to reach a destination, but~also to obtain features that improve pose estimation, map quality, spatial coverage, or~loop closure~\cite{leung2006active,placed2022explorb,bonetto2022irotate}. Visual search and active recognition acquire additional views because the location, identity, or~relevance of a target remains uncertain~\cite{laporte2006efficient,taioli2024unsupervised,bajpai2025uncertainty}. Active tracking faces a different source of uncertainty because the target and the observer may both move. Camera orientation, platform motion, or~camera assignment must therefore be adjusted to preserve visibility and avoid target loss~\cite{murray1994motion,rivlin2000control,casao2024distributed}. Robotic manipulation uses observation control when successful execution depends on evidence about object pose, grasp regions, contact, free space, or~task progress. Collaborative perception extends the same reasoning to multiple agents by deciding how observation, processing, and~communication responsibilities should be divided. Although~these tasks pursue different outcomes, they are included for the same reason. In~each case, the~system selects an action because of the evidence that the action is expected to make~available.

\subsection{Boundary with Neighboring Research~Areas}

The same criterion also clarifies the boundary between active visual perception and several neighboring fields. Conventional image and video analysis begins with an observation that has already been acquired. Detection, segmentation, recognition, and~visual question answering may improve how that observation is interpreted, but~they remain outside the present scope when they do not influence later data acquisition. Attention mechanisms do not change this distinction. Selecting image features or regions inside a fixed input changes internal processing, whereas active visual perception changes the evidence obtained from the environment. Offline multi view processing is excluded for the same reason when all views are captured according to a predetermined procedure. The~availability of several images may improve recognition or reconstruction, but~active selection is absent unless the system determines which additional view should be~acquired.

\begin{figure}[H]
\includegraphics[width=\linewidth]{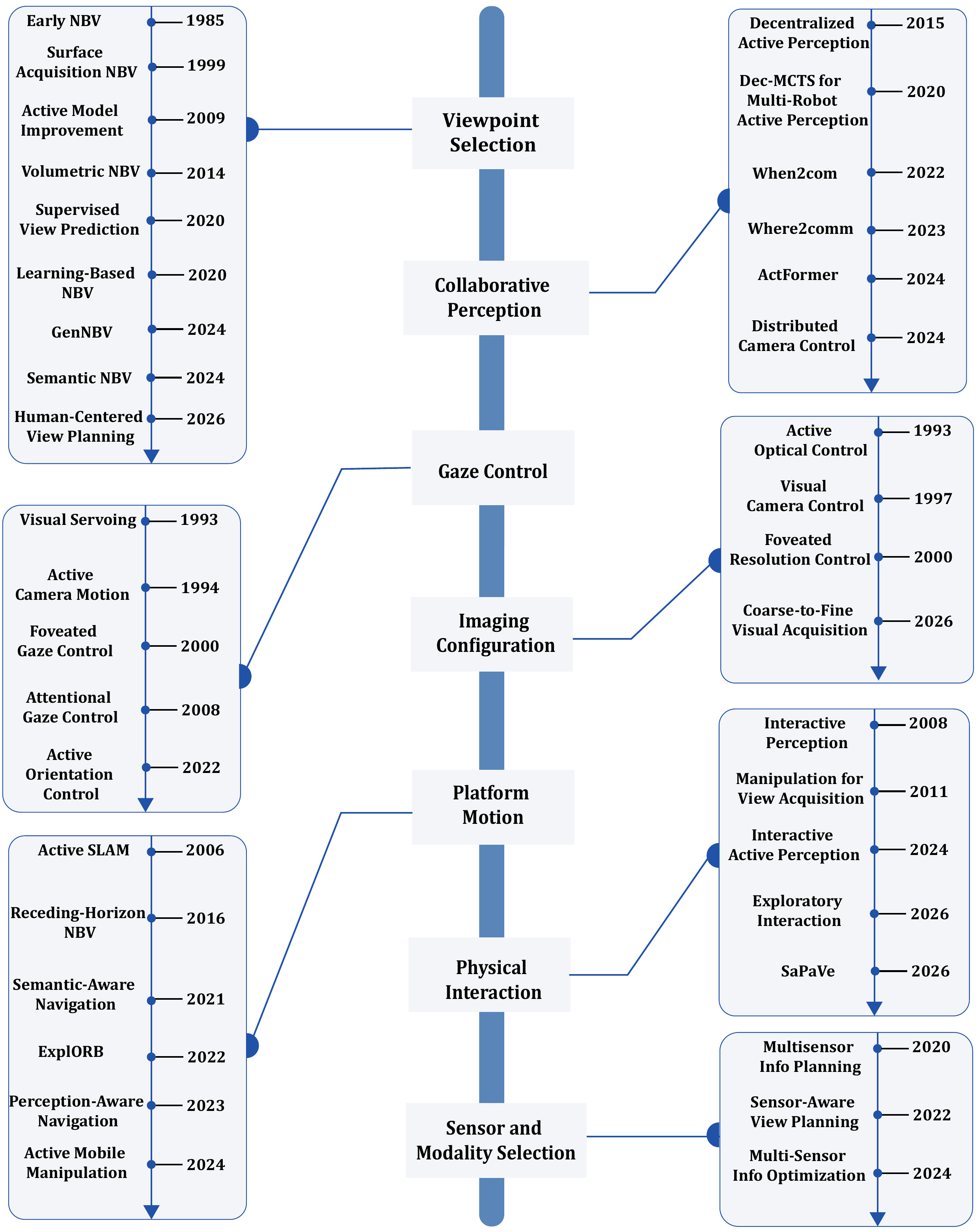}
\caption{\hl{Representative} 
 milestones in active visual perception organized by observation control~mechanism.}
\label{fig:timeline}
\end{figure}

Active learning also involves the selection of information, but~it operates at a different stage. It usually chooses existing samples for annotation or model training, while active visual perception chooses sensing or embodied actions that alter future observations. Other neighboring fields overlap more directly with the present scope. Active sensing includes decisions about modalities beyond vision and is therefore broader than active visual perception. Embodied perception studies how physical movement and interaction shape perception, although~it does not always require the agent to choose an action for the purpose of obtaining information~\cite{liu2025embodied}. Perception aware planning incorporates visibility, observability, or~feature quality into motion planning and becomes part of active visual perception when these considerations directly affect the acquisition of visual evidence. Multimodal fusion becomes active only when the system selects the sensor, observation, sampling pattern, or~transmitted information rather than combining a fixed set of inputs. These boundaries are not determined by terminology alone. A~study is included when its selected action changes the acquisition or allocation of visual evidence and when the resulting evidence contributes to a later perceptual or task decision. Figure~\ref{fig:timeline} highlights representative developments in observation control, from~camera and gaze control to platform motion, physical interaction, adaptive sensing, and~collaborative~perception.

\subsection{Literature Coverage and Study~Selection}
\label{sec:literature_selection}

The evidence base was developed through iterative searches in Google Scholar and Web of Science and was last updated on 5 September 2026. The~search strategy was designed to identify representative studies that provide broad coverage of the main forms of observation control and decision-making in active visual perception, rather than to support an exhaustive systematic review. The~search began with the broad terms ``active vision,'' ``active visual perception,'' and ``active sensing.'' These terms were then combined with task-specific and action-specific expressions, including ``next best view,'' ``active 3D reconstruction,'' ``active SLAM,'' ``active exploration,'' ``active visual search,'' ``active object recognition,'' ``active camera control,'' ``gaze control,'' ``active tracking,'' ``interactive perception,'' ``active perception for robotic manipulation,'' ``collaborative active perception,'' ``multi-agent active perception,'' and ``active sensor selection.'' Backward and forward citation tracing from foundational reviews and representative recent studies was used to identify influential earlier work, alternative terminology, and~studies addressing multiple task settings or forms of observation~control.

A study was included when three conditions were satisfied. First, visual evidence was required for the current perceptual estimate or task decision. Second, the~system selected a sensing or embodied action partly because of its expected effect on subsequent visual evidence. Third, the~newly acquired observation contributed to a later estimate, plan, communication decision, or~task action. The~selected action could control the viewpoint, gaze direction, camera configuration, platform motion, temporal or spatial sampling, sensing modality, physical interaction, sensor assignment, or~communication of visual~evidence.

Studies were excluded when they only processed fixed images or prerecorded videos, followed a predetermined trajectory without adapting to the perceptual state, combined a fixed set of sensor inputs, or~selected existing samples for annotation or model training. General motion planning was also excluded when visibility, observability, or~visual information quality did not influence the selected action. Offline multiview reconstruction was retained only when the system determined which additional observation should be acquired. These criteria apply the functional definition in Section~\ref{sec:definitions_scope} consistently across reconstruction, localization, search, tracking, manipulation, and~collaborative perception. The~main literature areas and representative search terms are summarized in Table~\ref{tab:search_terms}.

\begin{table}[H]
\caption{\hl{Literature} 
 areas and representative search terms used in this~review.}
\label{tab:search_terms}

\renewcommand{\arraystretch}{1.20}
\setlength{\tabcolsep}{4pt}
\footnotesize

\begin{tabularx}{\linewidth}{@{}
>{\hsize=0.86\hsize\linewidth=\hsize
  \raggedright\arraybackslash}X
>{\hsize=1.14\hsize\linewidth=\hsize
  \raggedright\arraybackslash}X
@{}}

\toprule
\textbf{Literature Area}
&
\textbf{Representative Search Terms}
\\
\midrule

Foundations and prior reviews
&
Active vision; active visual perception; active sensing;
revisiting active perception
\\
\midrule

Next best view and active reconstruction
&
Next best view; active 3D reconstruction; active view planning;
viewpoint selection for reconstruction
\\
\midrule

Active SLAM and exploration
&
Active SLAM; active visual SLAM; information gain for active SLAM;
semantic active SLAM; active exploration
\\
\midrule

Visual search, recognition, and~navigation
&
Active visual search; embodied visual search;
active object recognition; active perception for object navigation
\\
\midrule

Camera control, gaze, and~tracking
&
Active camera control; pan tilt zoom camera tracking;
gaze control; active viewpoint tracking; multi-camera tracking
\\
\midrule

Robotic manipulation and interactive perception
&
Active perception for robotic manipulation;
manipulation under occlusion; interactive perception;
active viewpoint selection for manipulation
\\
\midrule

Collaborative perception and sensor selection
&
Collaborative active perception; multi-agent active perception;
multi-robot exploration; cooperative next best view;
active sensor selection
\\

\bottomrule
\end{tabularx}

\noindent{\small{\hl{Note:} 
 The listed terms represent the main search directions used
to identify relevant literature rather than an exhaustive set of fixed
search queries. Additional studies were identified through backward and
forward citation tracing from foundational reviews and representative
recent work.}}
\end{table}
\vspace{-18pt}
\section{A Taxonomy from Perceptual Need to Sensing~Action}
\label{sec:taxonomy}

Active visual perception begins when the current observation cannot provide sufficient evidence for the next perceptual or task decision. The~missing evidence may concern an object hidden by occlusion, a~surface that has not been reconstructed, an~uncertain camera pose, a~target that is about to leave the field of view, or~a task region that cannot be inspected from the current position. In~distributed systems, the~required evidence may already be available to another sensor or agent but remain inaccessible because it has not been requested or transmitted. The~system must therefore determine what evidence is needed, how the observation can be changed, whether the expected improvement is worth the associated cost, and~which sensing action should be~executed.

These questions describe successive stages of the active visual perception loop. The~task first determines which part of the current perceptual state remains incomplete or unreliable. The~system then identifies the sensing actions that may expose the required evidence. Candidate observations are evaluated according to their expected contribution to perception and task execution, together with the motion, computation, communication, energy, and~safety costs required to acquire them. A~decision mechanism finally selects an action and updates the state after the new observation becomes available. The~effectiveness of an active perception method depends on all of these stages because an informative objective is of little use when the required action is unreachable, while a powerful decision model cannot compensate for a state representation that fails to describe the information needed by the~task.

\subsection{Perceptual Evidence Required by the~Task}

The first question in active visual perception concerns the evidence that is missing from the current observation. This question must be answered before a useful sensing action can be selected because different forms of perceptual insufficiency require different responses. A~target that is hidden behind an obstacle may require a change in viewpoint, while a distant target may require a narrower field of view or higher spatial resolution. An~uncertain object identity may require an observation that reveals a discriminative surface, whereas an uncertain robot pose may require visible landmarks or a loop closure. The~required evidence is therefore determined jointly by the current perceptual state and the task that the system is attempting to~complete.

\subsubsection{Occluded and Unobserved~Content}

Occlusion and incomplete scene coverage are among the most direct reasons for acquiring a new observation. In~active reconstruction, an~incomplete model indicates that some surfaces have not been observed or have been reconstructed with insufficient geometric support. The~system therefore searches for a viewpoint that exposes these regions and provides useful depth or appearance measurements~\cite{dunn2009next,vasquez2014volumetric,peralta2020next,chen2024gennbv,feng2024naruto,chen2025activegamer}. The~same problem appears in robotic manipulation when an object, grasp region, or~contact area is hidden by clutter. Changing the camera position may reveal the required region, but~some occlusions cannot be removed through camera motion alone because every reachable viewpoint remains~blocked.

Visibility also changes over time. In~active tracking, a~target may approach the image boundary, pass behind another object, or~become difficult to distinguish from the background. The~system must anticipate this loss of evidence and redirect the camera or platform before the target disappears~\cite{murray1994motion,rivlin2000control,casao2024distributed}. Visual search faces a related problem because the target may lie outside the current field of view or inside a region that has not yet been inspected~\cite{forssen2008informed,laporte2006efficient,taioli2024unsupervised}. Although~these tasks have different outputs, they share the need to expose visual content that is currently~unavailable.

\subsubsection{Uncertain State and Task~Progress}

The current observation may contain the relevant object or region but still fail to support a reliable decision. Active recognition methods acquire additional views when several object hypotheses remain consistent with the current image~\cite{roy2004active,laporte2006efficient}. Active localization and mapping methods seek observations that reduce uncertainty in the robot pose, map structure, or~data association~\cite{leung2006active,placed2022explorb,ahmed2023active,placed2023survey}.  In~these settings, the~missing evidence is not simply an invisible region. It is the information needed to distinguish between competing estimates of the current~state.

Task execution introduces further information requirements. A~manipulation system may need evidence about object pose, grasp stability, contact, free space, or~task progress rather than a complete description of the scene. A~navigation system may need to identify a traversable route or confirm a semantic landmark. Recent vision language systems extend these requirements to open vocabulary objects, spatial relations, and~instruction dependent regions~\cite{bajpai2025uncertainty,xue2026profocus,zhang2026embodied}. Their sensing decisions must still determine whether the current observation supports the requested conclusion or whether additional visual evidence is necessary. Semantic plausibility alone cannot replace observation because a model may produce a confident interpretation without having viewed the relevant~region.

\subsubsection{Complementary Evidence from Sensors and~Agents}

Within visually grounded active perception, another sensing modality may provide complementary evidence when the current visual observation is insufficient. A~visible light camera can provide texture and semantic detail, while depth sensors and LiDAR can provide geometric structure. Thermal sensing can reveal people and heat patterns under poor illumination, and~radar can remain useful when visual conditions deteriorate. The~need for an additional modality arises when the current sensor cannot reliably measure the property required by the task. Reflective surfaces, darkness, adverse weather, long distance, and~rapid motion can all change which sensing modality provides the most useful complementary evidence for the visually grounded~task.

A similar situation occurs in collaborative perception. Another vehicle, robot, or~camera may already have an unobstructed view of a region that is hidden from the current agent. The~perceptual need can then be addressed through observation assignment or communication rather than additional movement. Collaborative driving systems select spatial regions, features, or~queries that complement the local observation~\cite{hu2022where2comm,yang2023how2comm,huang2024actformer}. Multi-camera systems can transfer target tracking between cameras, while multi-robot systems can divide unexplored areas or direct different agents toward complementary viewpoints~\cite{atanasov2015decentralized,queralta2020collaborative,casao2024distributed}. The~useful evidence is determined by what the current system lacks, not by the total amount of information available across the~network.

\subsection{Actions That Change Visual~Observation}

Once the required evidence has been identified, the~system must determine how the observation can be changed. Some information can be obtained by redirecting the camera, while other information requires platform motion, a~change in imaging configuration, physical interaction, or~assistance from another agent. These actions differ in their physical effect, execution cost, and~degree of risk. They also determine which future observations are reachable and therefore constrain the decisions that can be made later in the closed~loop.

\subsubsection{Viewpoint, Gaze, and~Platform~Motion}

Changing the geometric relationship between the sensor and the scene is the most common way to acquire new evidence. The~system may change the camera position, orientation, viewing distance, or~platform pose. Classical next best view methods evaluate candidate camera poses according to visibility, surface coverage, or~uncertainty~\cite{connolly1985determination,pito1999solution,dunn2009next,vasquez2014volumetric}. Their decisions are interpretable because each candidate corresponds to a concrete geometric configuration. However, the~result depends strongly on how candidate poses are generated. A~sparse candidate set can exclude useful views, while a dense set increases the cost of visibility prediction, collision checking, and~motion~planning.

Learning-based methods predict a viewpoint or its expected value from the accumulated observation state~\cite{mendoza2020supervised,peralta2020next,chen2024gennbv}. They can represent complex relations between partial geometry and future coverage without evaluating every possible view online. They can also move beyond manually designed camera orbits. {One hybrid pipeline first acquires a single next-best view and then predicts a set of subsequent views with a multiview-conditioned network trained through long-tail sampling~\cite{WOS:001377255400009}.} Their predictions nevertheless depend on the geometry, sensor characteristics, and~action constraints represented during training. A~predicted viewpoint may appear informative in image space but remain unreachable or unsafe for the physical~platform.

The value of a viewpoint also changes with the task. Human centered view planning seeks observations that preserve useful views of a person rather than maximizing generic scene coverage~\cite{hu2026oa}. Sensor aware inspection considers whether a surface will produce a valid measurement rather than only whether it falls inside the field of view~\cite{yang2022next}. Plant inspection and semantic reconstruction prioritize structures or object classes that matter to the task~\cite{burusa2024gradient,yu2024semantic}. These studies show that geometric visibility is only the starting point. A~useful viewpoint must reveal the right evidence at a quality that supports the intended~decision.

Visual servoing provides a classical feedback mechanism for coupling visual measurements with camera or robot motion~\cite{hutchinson1996tutorial}. Gaze control changes the direction of observation without necessarily changing the camera position~\cite{vogel2008target}. A~pan and tilt camera can follow a moving target while the platform remains stationary, and~a mobile robot can rotate its camera or body toward useful landmarks without changing its translational route. Camera motion can preserve target visibility, but~it also produces background motion and control error that must be separated from target motion~\cite{papanikolopoulos1993visual,murray1994motion}. Foveated systems add another decision because high resolution sensing is concentrated near the fixation point, while the periphery is observed at lower resolution~\cite{rivlin2000control}. Active localization methods also use orientation control to observe stable or informative features even when the route remains unchanged~\cite{frintrop2008attentional,bonetto2022irotate}.

Platform motion extends viewpoint control because the same movement changes both the visual observation and the physical state of the task~\cite{velez2011planning}. Active localization and mapping methods move toward unexplored regions, informative landmarks, or~potential loop closures~\cite{leung2006active,placed2022explorb}. Perception aware navigation favors routes that maintain stable visual features or semantically useful regions~\cite{bartolomei2021semantic,wang2023active}. Mobile manipulation must further consider whether a better observation justifies additional travel, collision risk, and~interruption of the main task~\cite{jauhri2024active}. A~camera pose that cannot be reached through a safe and efficient trajectory has limited practical value even when its predicted visual quality is~high.

\subsubsection{Imaging Configuration and Adaptive~Sampling}

Visual evidence can also be changed without moving the camera. Focus, zoom, aperture, exposure, field of view, spatial resolution, frame rate, and~temporal sampling determine the quality and frequency of the measurements delivered to the perception system. Early active vision research treated these controls as part of the sensing process because a different viewpoint cannot correct blur, saturation, poor scale, or~limited depth of field~\cite{swain1993promising,nelson1997visually}. Recent manipulation systems have renewed interest in coarse to fine sensing, where a broad observation identifies a relevant region and a focused observation supports precise action~\cite{liu2026activevla}.

Optical and temporal controls are closely coupled to perception failures. Zoom can provide more detail but narrows the field of view and increases the risk of target loss. A~longer exposure can improve low light appearance but also increase motion blur. A~higher frame rate can improve tracking of rapid motion but consumes additional computation, bandwidth, and~energy. The~same sensing configuration is therefore not optimal throughout a task. A~system may begin with wide and inexpensive observations, allocate greater resolution when uncertainty becomes concentrated in a small region, and~reduce the sampling rate when the scene changes~slowly.

These controls remain less developed than viewpoint planning. Many systems mention adjustable resolution or zoom but treat them as fixed hardware settings rather than actions selected from the current perceptual state. Most visual benchmarks also assume a fixed frame rate and fixed image quality. This assumption hides an important capability of physical systems because changing the sensing configuration may sometimes provide more useful evidence at a lower cost than moving the entire~platform.

\subsubsection{Physical~Interaction}

Camera and platform motion cannot reveal all task-relevant information. In~cluttered environments, every collision free viewpoint may remain occluded, and~some object properties only become observable after the object has been moved or touched. Interactive perception addresses this limitation by changing the observed object or the surrounding environment. A~robot may push an occluder, rotate an object, open a container, grasp and reorient a part, or~establish contact that reveals material or mechanical properties~\cite{le2008active,krainin2011autonomous,weng2024interactive,jiang2024roboexp}.

Interaction can expose evidence unavailable through visual observation alone. Repositioning an object can reveal a hidden surface and improve pose estimation. Interaction-driven reconstruction extends this idea to object interiors by manipulating articulated parts to expose geometry that cannot be scanned from external viewpoints alone~\cite{WOS:001139790400078}. Repeated contact can reveal articulation, deformability, mass distribution, or~tool function. {Predictive visuo-tactile perception selects informative pushing and pulling actions to infer indirectly observable physical properties, including mass, center of mass, and~friction~\cite{WOS:001421981700003}.} Recent systems coordinate exploratory actions with task directed behavior so that the robot first reduces uncertainty and then performs the intended manipulation~\cite{he2026towards,weng2024interactive}. The~sensing action and the task action may therefore become part of the same process rather than two separate~stages.

The additional information comes with greater physical risk. An~interaction may damage an object, move it into a less observable configuration, create an irreversible state, or~interfere with later task execution. Contact outcomes are also difficult to predict accurately. Interactive perception is therefore most appropriate when the required evidence cannot be obtained safely through camera or platform motion and when the expected information justifies the possibility of disturbing the~scene.

\subsubsection{Collaborative Observation and~Communication}

Multiple agents can enlarge the observable region and reduce individual blind spots, but~their observations become useful only when sensing responsibilities are coordinated. Multi-robot exploration assigns different regions, frontiers, or~observation tasks to different agents and coordinates their motion to improve exploration efficiency and sensing coverage~\cite{zhou2023racer,chiun2025marvel}. Multi-camera tracking determines which camera should observe a target and when another camera should take over. Collaborative driving allows vehicles and roadside units to provide evidence about regions that are occluded from the local~vehicle.

The action in these systems may involve movement, sensor assignment, message selection, or~communication timing. Classical decentralized methods seek informative joint behavior without relying entirely on a central controller~\cite{atanasov2015decentralized,queralta2020collaborative}. Recent collaborative perception systems select spatial regions, features, or~queries rather than transmitting all available data~\cite{hu2022where2comm,yang2023how2comm,huang2024actformer}. Distributed camera control coordinates viewpoints so that target visibility and identity can be maintained across a wider area~\cite{casao2024distributed}.

The benefit of collaboration comes from complementary evidence rather than the number of sensors alone. Repeated transmission of information already available to the receiving agent wastes bandwidth and processing resources. Coordination becomes more difficult when agents have different sensing capabilities, uncertain relative poses, delayed messages, or~inconsistent local maps. The~system must therefore determine who should observe, what should be transmitted, whether the information will remain valid when it arrives, and~how perception should continue when communication becomes~unreliable.

\subsection{Value of the Acquired~Observation}

Changing the observation is useful only when the newly acquired evidence improves perception or task execution. The~value of an observation cannot be determined from visual change alone because a different image may contain no additional task-relevant information. Active perception systems therefore predict how a candidate observation will affect visibility, uncertainty, geometric quality, task outcome, or~resource consumption. Different objectives emphasize different aspects of this improvement, and~their suitability depends on what the system currently needs to~know.

\subsubsection{Visibility, Coverage, and~Geometric~Quality}

Visibility provides the most direct measure of whether a sensing action reveals the required target, surface, or~region. A~tracking system can seek to keep a target inside the field of view, a~manipulation system can seek an unobstructed view of a grasp region, and~a reconstruction system can seek viewpoints that expose unobserved surfaces. Coverage extends this reasoning from a single target to the amount of scene or object geometry observed by the system. Surface completeness, valid depth, map coverage, and~frontier reduction are common measures in reconstruction and~exploration.

These objectives are easy to interpret, but~visibility alone does not guarantee useful evidence. A~surface may be visible but too distant, blurred, reflective, or~poorly illuminated for reliable measurement. A~person may occupy a large image region but appear from an angle that does not support pose estimation or identification. A~large unknown region may increase map coverage while contributing little to the current task. Sensor aware and target-oriented methods address these limitations by combining geometric visibility with measurement validity, semantic relevance, or~task affordance~\cite{yang2022next,zhang2023affordance,yu2024semantic,hu2026oa}.

Coverage can also conflict with localization quality. An~exploration policy may repeatedly enter unknown areas even as pose uncertainty increases. Conversely, a~localization-oriented policy may remain near familiar landmarks and fail to expand the map. Active localization and mapping methods therefore combine exploration value with feature quality, trajectory cost, and~loop closure potential~\cite{placed2022explorb,wang2023active}. The~central question is not how much new space can be observed, but~whether the added coverage can be obtained while maintaining a reliable perceptual~state.

\subsubsection{Information Gain and Uncertainty~Reduction}

Many active perception methods value an action according to the expected reduction in uncertainty. The~uncertain state may describe object identity, surface occupancy, robot pose, map structure, target location, or~semantic belief. The~system predicts how a candidate observation will change this state and prefers observations expected to distinguish competing hypotheses or concentrate the posterior~estimate.

This principle appears across different tasks. Active recognition selects
views that separate object hypotheses
~\cite{laporte2006efficient}.
{Active reconstruction values observations according to uncertainty over
occupancy, surface geometry, and~radiance field representations, while
hierarchical uncertainty can be used to guide next best viewpoint selection}
\cite{WOS:001589384100003}.
Active localization and mapping use pose covariance, map uncertainty, or~information matrices to estimate the value of candidate movements
~\cite{mu2016information,ahmed2023active,carrillo2012comparison}.
Search systems use spatial or semantic beliefs to determine which region is
most informative to inspect next. The~mathematical form may be shared, but~the
meaning of uncertainty depends on the state representation used by the~system.

The quality of an information driven decision depends on the calibration of that representation~\cite{rodriguez2018importance}. A~system can be confidently wrong when the scene differs from the training data or when the observation model ignores sensor failure. A~geometric model may overestimate the value of a reflective surface, while a vision language model may produce high semantic confidence without sufficient visual support. Inaccurate uncertainty can cause the system to seek the wrong evidence and reinforce an incorrect belief. Evaluation should therefore compare predicted observation value with the improvement that is actually obtained after the action, rather than reporting only the final task~result.

\subsubsection{Downstream Task~Utility}

{Task utility provides active perception with a task-specific evaluation criterion. Its core principle is that a sensing action should be selected according to whether the resulting observation can improve subsequent decisions, rather than merely increase visibility or reduce general uncertainty} \cite{WOS:001297903800001,WOS:001329156500001,WOS:001848650500001}. {This principle takes different forms across tasks. In~manipulation, a~useful viewpoint provides evidence that improves grasping, insertion, contact verification, or~gripper pose selection} \cite{WOS:001434749300008,WOS:001574207700008,WOS:001694936300008,WOS:001676224400011}. {In navigation, active target perception uses real time detection feedback and task constraints to adjust the robot position and provide a suitable viewpoint for subsequent manipulation} \cite{qin2025active,WOS:001717837400001,WOS:001262727500008,WOS:001689935800001}. {In visual search, next best view planning directs the sensor toward viewpoints that reveal occluded targets and improves search decisions under uncertain detections} \cite{WOS:001638088100005,taioli2024unsupervised,WOS:001797477500001}. {In collaborative perception, communication is useful when shared information fills missing perceptual evidence and consequently improves the receiving agent's decisions} \cite{hu2024communication,WOS:001587374100001,WOS:001360457500036,WOS:001173317800107}. {These examples demonstrate that the value of an observation depends on how the evidence it provides influences task execution.} \par {However, the~final task outcome alone does not reveal whether an improvement results from better observations or from other components of the policy. A~reinforcement learning policy may achieve a higher success rate because it has learned favorable motion patterns rather than because it acquires better visual evidence. A~vision language action model may also benefit from stronger semantic or action priors even when active camera control contributes little. Controlled comparisons between active and fixed camera configurations can help isolate the contribution of viewpoint adaptation} \cite{xiongvision,WOS:001342557800026}. {Evaluation should therefore report observation quality, sensing decisions, controller performance, and~final task outcomes separately} \cite{WOS:001566941700012}. \par {This distinction is especially important in recent manipulation systems that tightly couple visual acquisition with physical action} \cite{weng2024interactive,WOS:001862921800014}. {The dual arm DAVMM model predicts eye and hand actions simultaneously for manipulation under occlusion} \cite{WOS:001454288700001}. {SaPaVe jointly optimizes semantic camera control and manipulation through separate but coordinated action spaces} \cite{liu2026sapave}, {while ActiveVLA uses coarse to fine region localization, active view selection, and~spatial refinement to support precise manipulation} \cite{liu2026activevla}. {In these systems, the~useful observation depends on the next physical action, while the action also depends on the evidence acquired from the selected viewpoint. Active perception therefore remains coupled to task execution throughout the closed loop.}

\subsubsection{Operational Cost and~Risk}

An observation that improves perception may still be unsuitable when the cost of acquiring it is too high. Camera motion, robot travel, sensing, model inference, communication, and~physical interaction all consume limited resources. These costs determine whether active perception improves the complete system rather than only the accuracy of an isolated perception~model.

Motion cost can be represented by path length, rotation, action count, execution time, or~interruption of the main task. Computational cost includes candidate generation, observation prediction, optimization, and~network inference. Energy is critical for UAVs, mobile robots, and~long-term monitoring systems. Communication cost affects collaborative perception because transmitting dense features or raw images can exceed available bandwidth. Risk includes collision, unstable viewpoints, object damage, privacy intrusion, and~actions that leave the system without a safe recovery~path.

Many methods combine expected benefit and cost within a weighted objective or constrained optimization. A~single combined score can simplify action selection but makes comparison difficult because the meaning of each weight changes across platforms. One meter of motion has a different cost for a ground vehicle, a~UAV, and~a wrist mounted camera. Reporting task improvement and individual costs separately provides a clearer account of whether the method remains useful when the resource budget or deployment platform~changes.

\subsection{Selection of the Next Sensing~Action}

After the system has defined the available actions and the expected value of their resulting observations, it must determine which action should be executed. The~difficulty of this decision depends on the number of candidate actions, the~uncertainty of the current state, the~availability of an observation model, and~the time over which future consequences must be considered. Simple and well structured tasks can often be handled through explicit rules or geometry, while uncertain and long-term tasks require probabilistic reasoning, planning, learning, or~combinations of several~mechanisms.

\subsubsection{Geometric and Rule-Based~Decisions}

Rule-based methods select sensing actions through designed priorities, thresholds, and~state transitions. A~camera may recenter a target when it approaches the image boundary. A~robot may rotate toward a landmark when the number of visible features falls below a threshold. A~search system may inspect salient regions before less relevant regions. These decisions are computationally efficient and easy to interpret when the operating conditions are well~defined.

Geometry-based methods use visibility, occlusion, projection, surface orientation, baseline, reachability, and~collision constraints to evaluate candidate actions~\cite{WOS:001162837000001}. They remain effective in reconstruction and inspection because the relation between camera pose and surface observation has a clear physical basis~\cite{dunn2009next,yang2022next}. They do not require large training datasets, and~their failures can often be traced to a specific geometric~assumption.

Their limitations become more apparent when the scene is dynamic, deformable, semantically ambiguous, or~difficult to model. A~geometric policy requires a sufficiently accurate partial map or object representation. It may also treat all visible surfaces as equally measurable even when sensor response depends on material, lighting, or~motion. Practical systems often preserve geometric constraints for safety and feasibility while using other mechanisms to estimate task relevance or observation~quality.

\subsubsection{Probabilistic and Information-Based~Decisions}

Probabilistic methods maintain a belief over uncertain states and predict how candidate observations will modify that belief. Bayesian active recognition selects views that distinguish object hypotheses~\cite{laporte2006efficient}. Active localization and mapping propagate pose and map uncertainty through possible controls. Belief-based reasoning is well suited to partial observability because the system can represent several possible states instead of committing immediately to one~estimate.

{Building on such belief representations, information-based active
perception compares sensing actions according to their expected reduction in
task-relevant uncertainty. Active SLAM methods implement this principle using
information gain or KL divergence to estimate how future observations may
reduce pose and map uncertainty}
\cite{mu2016information,carlone2014active}.
{Recent work extends this formulation to dynamic semantic environments by
combining mutual information with hierarchical NBV decisions to improve path
planning, mapping efficiency, and~tracking robustness}
\cite{WOS:001688974600025}.
{Fisher information applies the same principle to active view selection in
radiance field mapping}
\cite{jiang2024fisherrf}.
{Submodular formulations support coordinated observation selection under
constraints involving multiple sensors or multiple robots}
\cite{lauri2020multi,bai2024multi}.
{These criteria provide a common framework for connecting sensing actions
with expected uncertainty reduction, but~their reliability depends on the
belief and observation models from which their scores are computed. Poorly
represented uncertainty or omitted sensor failure modes may therefore cause a
mathematically optimal action to provide little useful perceptual evidence}
\cite{rodriguez2018importance}.
{This dependence becomes more consequential as the belief expands to
include several coupled sources of uncertainty.}

A realistic belief may need to describe robot pose, map uncertainty, target identity, object motion, sensor reliability, and~task state at the same time. Exact prediction becomes difficult as the state and action spaces grow. Practical methods therefore simplify the belief, approximate future observations, restrict candidate actions, or~consider only a limited future horizon. These approximations reduce computation but must be reported clearly because they determine which uncertainty sources influence the final~decision.

\subsubsection{Search and Receding Horizon~Planning}

A single sensing action may have consequences that only become visible after several later actions. A~locally informative view can move the system into a poor position for subsequent observation, while an initially less informative action can create access to a sequence of valuable views. Search and planning methods address this dependence by evaluating candidate sequences, graph paths, frontiers, or~trajectories~\cite{yan2023active,popovic2024learning,bircher2016receding,atanasov2014nonmyopic}.

Graph search- and sampling-based methods can compare discrete viewpoints and motion sequences. {For multi-target active tracking, Monte Carlo tree search can be combined with target-motion prediction to select camera actions~\cite{chen2024optimizing}.} Trajectory optimization can handle continuous platform or camera motion. Model predictive control evaluates actions over a limited future horizon, executes the first action, and~replans after the next observation~\cite{avni2008scanning,leung2006active,chen2020active,wang2023active,allibert2010predictive}. This repeated update preserves responsiveness to newly acquired evidence and reduces the computational burden of full long-term~planning.

Planning quality depends on both the candidate generator and the model used to predict future observations. A~planner cannot select a viewpoint that is absent from the action space. It can also overestimate a candidate when the prediction ignores blur, calibration error, reflective surfaces, dynamic occlusion, or~communication delay. The~planning mechanism must therefore be evaluated together with the state representation and observation model on which its predictions~depend.

\subsubsection{Learning-Based~Policies}

Supervised methods learn a mapping from the current observation state to a next viewpoint or an estimate of viewpoint utility~\cite{mendoza2020supervised,chen2024gennbv}. By~learning from labelled viewpoints or supervision derived from view quality, these methods can reduce the need for exhaustive online evaluation of candidate observations. Imitation learning and hybrid behavior cloning and reinforcement learning approaches can acquire active sensing strategies from demonstrations and interaction, including visual search, tracking, focusing, and~task coupled gaze control~\cite{xiongvision,kerr2025eye,WOS:001542495900011}. Both paradigms enable efficient policy inference, but~their performance depends on how well the training data represent the observation states, sensing actions, and~task conditions encountered at~deployment.

Reinforcement learning learns sensing policies through interaction and can account for rewards that appear only after several actions. It has been applied to reconstruction, active localization and mapping, semantic navigation, tracking, robotic manipulation, and~camera networks~\cite{cheng2018reinforcement,peralta2020next,placed2020deep,bartolomei2021semantic,li2020pose,huang2023earl}. A~reward can combine perceptual improvement, task success, and~action cost, allowing the policy to adapt its behavior to long-term~consequences.

The main difficulty is transfer. A~policy may exploit visual appearance, motion dynamics, action discretization, or~reward structure specific to the training environment. Changes in camera calibration, actuator delay, lighting, object dynamics, or~sensor noise can invalidate the learned relation between action and observation. Evaluation should therefore include unseen environments, perturbations, and~real platforms rather than only new episodes generated from the same simulator~\cite{hu2026real}.

\subsubsection{Semantic Reasoning with Foundation~Models}

Foundation models provide language conditioned representations and broad semantic knowledge that can help identify which objects, regions, or~relations are relevant to a task. Open vocabulary navigation methods use uncertainty or semantic relevance to guide exploration~\cite{bajpai2025uncertainty}. ProFocus generates targeted visual queries when the current evidence is insufficient~\cite{xue2026profocus}. Embodied Reasoner connects observation, reasoning, and~action over embodied trajectories~\cite{zhang2026embodied}. ActiveVLA and SaPaVe further connect semantic reasoning with camera control and manipulation~\cite{liu2026activevla,liu2026sapave}.

These systems extend active perception beyond fixed categories and predefined task states. However, semantic confidence is not equivalent to visual evidence. A~model may generate a plausible conclusion even when the relevant object or region has not been observed. The~sensing process must distinguish prior semantic knowledge from evidence acquired in the current environment and must verify that a proposed viewpoint is reachable, safe, and~likely to resolve the~uncertainty.

Latency and reproducibility also affect closed-loop use. A~large model may delay sensing decisions, and~external model services may change over time. Reliable evaluation requires reporting the exact model, visual input resolution, prompt, decoding settings, reasoning interface, and~action representation. Comparisons with simpler semantic, geometric, and~learning-based baselines are also necessary to determine whether the foundation model improves observation selection rather than only providing a stronger perception~backbone.

\subsubsection{Hybrid and Distributed Decision~Making}

Many physical systems combine several decision mechanisms because no single method addresses semantic relevance, uncertainty, feasibility, long-term effects, and~communication at the same time. {A dual-resolution NBV planner for fusion reactor maintenance separates global low-resolution subregion selection from local high-resolution viewpoint selection, connecting large-scale mapping decisions with detailed observation planning~\cite{WOS:001785709100001}.} A semantic model may identify a task-relevant region, a~geometric module may generate reachable viewpoints, an~information objective may rank them, and~a motion planner may connect the selected view to the current platform state. A~learned policy may estimate long-term value, while a distributed protocol determines which evidence should be shared. Semantic next best view planning, active localization and mapping, collaborative perception, and~active manipulation commonly follow this form~\cite{yu2024semantic,placed2022explorb,hu2022where2comm,liu2026sapave}.

Hybrid systems can combine the transparency of explicit models with the flexibility of learned representations. Their main difficulty is attribution because several components may change simultaneously. An~apparent improvement in active perception may result from a stronger perception backbone, a~larger action space, a~more accurate state representation, or~a better controller rather than the sensing policy itself. Module level ablations and intermediate measurements are therefore necessary to show which part of the system improves the acquired evidence and which part improves the downstream~task.

Distributed systems add another dependency because each agent acts on incomplete local information. The~usefulness of a sensing action can depend on what other agents have observed, which messages have arrived, and~whether shared information remains temporally valid. A~robust distributed policy must continue operating under delayed or missing communication and should avoid relying on globally synchronized states that are unavailable during real deployment. {The interpretation of these limitations also depends on the evaluation environment. Most representative methods summarized in} Table~\ref{tab:decision_mechanisms} {were evaluated primarily in simulated or benchmark environments, while physical-platform validation was reported only for a limited subset. The~limitations supported by these evaluations therefore mainly concern computational scalability, model assumptions, training-data coverage, reward design, and~simulated communication or interaction conditions. Potential sensitivity to calibration errors, sensor noise, control delay, dynamic occlusion, communication loss, reachability constraints, and~safety risks should consequently be regarded as insufficiently validated deployment concerns unless they have been examined on physical platforms. The~table therefore distinguishes limitations supported by the reported evaluation settings from the status of physical-platform validation.}

\begin{table}[H]
\caption{\hl{Comparison} 
 of representative mechanisms for selecting sensing actions in active visual perception. The~reported simulation limitations and the status of physical-platform validation are~distinguished.}
\label{tab:decision_mechanisms}

\begin{adjustwidth}{-\extralength}{0cm}
\centering
\renewcommand{\arraystretch}{1.08}
\setlength{\tabcolsep}{2.5pt}
\scriptsize

\begin{tabularx}{\fulllength}{@{}
    >{\hsize=0.85\hsize\linewidth=\hsize
      \raggedright\arraybackslash}X
    >{\hsize=0.80\hsize\linewidth=\hsize
      \raggedright\arraybackslash}X
    >{\hsize=0.85\hsize\linewidth=\hsize
      \raggedright\arraybackslash}X
    >{\hsize=1.50\hsize\linewidth=\hsize
      \raggedright\arraybackslash}X
@{}}

\toprule
\textbf{Decision Mechanism and Representative Studies}
&
\textbf{Required Information}
&
\textbf{Suitable Sensing Decisions}
&
\textbf{Evaluation Setting and Main Limitation}
\\
\midrule

Rule-based and geometric
~\cite{frintrop2008attentional,rivlin2000control,
dunn2009next,yang2022next}
&
Designed states, geometric models, visibility, and~reachability
&
Gaze control, camera pose, viewpoint, and~platform motion
&
\textit{\hl{Simulation evidence.} 
} Evaluation is limited to simulation,
where performance depends on designed rules, geometric accuracy,
and idealized visibility.
\textit{\hl{Physical-platform validation.}} Not reported; robustness to
calibration errors, dynamic occlusion, material effects, and~actuation
uncertainty remains unverified.
\\

\midrule

Probabilistic and belief-based
~\cite{laporte2006efficient,kim2012next,
kitanov2024topological}
&
State beliefs, observation likelihoods, and~uncertainty models
&
Viewpoint, revisit action, path, and~sensing sequence
&
\textit{\hl{Simulation evidence.}} Evaluation is limited to simulation,
where computation increases with belief dimensionality and planning
horizon and depends on assumed noise models.
\textit{\hl{Physical-platform validation.}} Not reported; robustness to
sensor failures, likelihood miscalibration, measurement noise, and~real-time constraints remains unverified.
\\

\midrule

Information-based optimization
~\cite{mu2016information,lauri2020multi,
bai2024multi}
&
Calibrated state uncertainty and predicted observation updates
&
Viewpoint, waypoint, sensor, and~measurement selection
&
\textit{\hl{Simulation evidence.}} Evaluation is limited to simulation,
where observation values inherit errors from idealized belief and
sensor models.
\textit{\hl{Physical-platform validation.}} Not reported; whether predicted
information gain is preserved under sensor noise, model mismatch, and~execution delay remains unverified.
\\

\midrule

Search and receding horizon planning
~\cite{leung2006active,chen2020active,
wang2023active,taioli2024unsupervised}
&
Candidate actions, transition models, and~predicted future observations
&
View sequences, exploration routes, and~continuous camera motion
&
\textit{\hl{Simulation evidence.}} Evaluation is limited to simulation,
where computation grows with the action space and prediction depth
and future observations are often idealized.
\textit{\hl{Physical-platform validation.}} Not reported; robustness to
planning latency, motion uncertainty, dynamic obstacles, and~reachability constraints remains unverified.
\\

\midrule

Supervised and imitation learning
~\cite{mendoza2020supervised,xiongvision,
chen2024gennbv}
&
View labels, expert values, or~demonstrations
&
Viewpoint, gaze, zoom, and~camera behavior
&
\textit{\hl{Simulation evidence.}} Evaluation is limited to simulation,
where policies inherit training-data coverage limits and may overfit
simulated appearance or action discretization.
\textit{\hl{Physical-platform validation.}} Not reported; robustness to
changes in lighting, camera calibration, motion blur, and~actuator
response remains unverified.
\\

\midrule

Reinforcement learning
~\cite{peralta2020next,placed2020deep,
bartolomei2021semantic,huang2023earl}
&
Interaction experience, rewards, and~environment dynamics
&
Path, viewpoint, gaze, and~task coupled sensing actions
&
\textit{\hl{Simulation evidence.}} Evaluation is limited to simulation,
where policies remain sensitive to reward design and may exploit
simulator-specific appearance or dynamics.
\textit{\hl{Physical-platform validation.}} Not reported; sim-to-real
transfer, safe exploration, control delay, and~reliability under
physical disturbances remain unverified.
\\

\midrule

Foundation model-based reasoning
~\cite{bajpai2025uncertainty,xue2026profocus,
zhang2026embodied,liu2026activevla}
&
Language instructions, semantic representations, and~reasoning history
&
Semantic queries, relevant regions, viewpoints, and~embodied actions
&
\textit{\hl{Simulation evidence.}} Evaluation is limited to simulated or
benchmark environments, where semantic confidence may be mistaken
for visual evidence and results remain sensitive to prompts and model
versions.
\textit{\hl{Physical-platform validation.}} Not reported; real-time visual
grounding, action reachability, inference latency, and~safe closed-loop
execution remain unverified.
\\

\midrule

Collaborative and distributed
~\cite{atanasov2015decentralized,hu2022where2comm,
huang2024actformer,casao2024distributed}
&
Local observations, agent states, message value, and~communication status
&
Agent assignment, viewpoint coordination, and~message selection
&
\textit{\hl{Simulation evidence.}} Evaluation is limited to simulation,
where communication, synchronization, and~pose information are
typically simplified.
\textit{\hl{Physical-platform validation.}} Not reported; robustness to
packet loss, variable delay, limited bandwidth, pose error, and~temporal misalignment remains unverified.
\\

\midrule

Interactive and hybrid
~\cite{le2008active,krainin2011autonomous,
weng2024interactive,yu2024semantic,
liu2026sapave}
&
Task state, physical constraints, semantic relevance, and~predicted outcomes
&
Physical interaction and combinations of sensing, motion,
and communication actions
&
\textit{\hl{Simulation evidence.}} Evaluation in simulated settings shows that module coupling and idealized interaction models can make performance attribution difficult.
\textit{\hl{Physical-platform validation.}} Reported in at least one representative study; physical robot experiments expose practical constraints including restricted camera workspaces and interaction uncertainty. Broader robustness to hardware variation, unrestricted environments, and~physical safety remains insufficiently validated.
\\

\bottomrule
\end{tabularx}

\end{adjustwidth}
\noindent{\small{\hl{Note.} ``Simulation evidence'' denotes limitations supported
by the reported simulation or benchmark experiments.
``Physical-platform validation not reported'' indicates that the
corresponding deployment concerns have not been experimentally
demonstrated on a physical real-world platform.}}

\end{table}
\vspace{-15pt}
\subsection{Cross-Task Regularities in Active Observation~Selection}

The preceding analysis reveals several regularities that recur across reconstruction, localization, search, tracking, manipulation, and~collaborative perception. These regularities do not depend on a particular task name or action selection algorithm. They arise from the relationship among the evidence required by the task, the~action used to acquire that evidence, the~predicted value of the resulting observation, and~the physical process through which the action is~executed.

First, observation value is meaningful only in relation to the state variable or task decision that requires additional evidence. Information gain is often presented as a general objective, but~the uncertain quantity differs substantially across applications. Active recognition seeks observations that distinguish competing object hypotheses~\cite{laporte2006efficient,roy2004active}. Active localization and mapping seek measurements that reduce uncertainty in the robot pose, map structure, or~data association~\cite{mu2016information,placed2022explorb,ahmed2023active}. Active reconstruction seeks evidence about unobserved or poorly supported surfaces~\cite{peralta2020next,chen2024gennbv}. Collaborative perception seeks information that complements the receiving agent rather than information that is merely uncertain at the transmitting agent~\cite{hu2022where2comm,yang2023how2comm}. A~numerical reduction in uncertainty therefore does not have a uniform interpretation. Its practical value depends on whether the reduced uncertainty concerns a variable that can change the next task~decision.

Second, the~action selection mechanism should not be confused with the objective of active perception. Reinforcement learning, probabilistic planning, geometric optimization, and~foundation model reasoning describe procedures for selecting an action. They do not by themselves specify what evidence the task requires or why a particular observation is useful. Reinforcement learning can optimize surface coverage, localization quality, search success, target visibility, manipulation success, or~communication efficiency according to its state and reward~\cite{peralta2020next,placed2020deep,bartolomei2021semantic,huang2023earl}. Foundation models can identify semantically relevant objects or regions, but~this semantic relevance must still be connected to an observable visual condition and a task decision~\cite{xue2026profocus,zhang2026embodied,liu2026activevla}. The~capability of the action selection mechanism should therefore be evaluated separately from the perceptual objective that it is expected to~serve.

Third, predicted observation value and physical executability are two different requirements. A~candidate viewpoint may provide high geometric coverage or semantic relevance while remaining unreachable because of joint limits, obstacles, platform dynamics, or~safety constraints. Sensor properties may also prevent a theoretically visible surface from producing a valid measurement~\cite{yang2022next}. This distinction is especially important in mobile manipulation, UAV inspection, and~human centered sensing, where the camera pose changes the physical state of the platform and may interfere with the following task action~\cite{jauhri2024active,hu2026oa}. A~method should therefore report both the predicted benefit of a candidate observation and the feasibility, time, risk, and~control error involved in acquiring it. Treating reachability as a final filtering step can lead the system to favor observations whose apparent value disappears after motion constraints are~considered.

Fourth, final task performance alone cannot demonstrate that active observation selection was effective. A~manipulation system may achieve higher success because it uses a stronger pose estimator or controller. A~navigation policy may benefit from improved semantic representations even when its sensing actions provide little additional evidence. A~collaborative system may improve detection by transmitting more features rather than selecting more complementary information. The~contribution of active perception becomes clearer when evaluation measures the information deficit before the action, the~direct change in observation quality after the action, the~resulting change in task performance, and~the resources used to obtain that change. Module level ablations are also required when the perception backbone, state representation, action space, controller, and~sensing policy change~simultaneously.

Fifth, fixed datasets cannot fully evaluate a closed-loop sensing policy when the policy is expected to change future observations. Prerecorded images and trajectories can test detection, depth estimation, localization, reconstruction, or~observation value prediction under controlled inputs. They cannot determine whether another action would have produced a more useful observation because every policy receives the same predetermined evidence. The~Active Vision Dataset provides a limited graph of selectable observations captured in real indoor environments~\cite{ammirato2017dataset}. Habitat and RoboTHOR allow an embodied policy to change its position and acquire different future views~\cite{savva2019habitat,deitke2020robothor}. Physical platforms remain necessary when the research question involves reachability, actuation error, camera vibration, physical contact, safety, or~communication delay. Closed-loop claims should therefore be supported by an environment in which the selected action actually changes the next~observation.

A further regularity concerns the decision to stop sensing. Additional observations are not automatically beneficial. Once the current evidence is sufficient to support the task decision, another view can add motion, latency, computation, energy consumption, or~physical risk without changing the outcome. Conversely, stopping too early leaves an unresolved information deficit. The~stopping condition should therefore depend on the expected effect of another observation on the task rather than only on a fixed number of actions. This requirement connects evidence sufficiency, observation value, acquisition cost, and~task progress within a single~decision.

Together, these regularities show that active visual perception cannot be established merely by demonstrating that another viewpoint contains information or that an active policy achieves a higher final score. A~convincing evaluation must show that the task required additional visual evidence, that the selected action could acquire that evidence under realistic constraints, that the new observation improved the relevant perceptual state, and~that the resulting task benefit justified the acquisition cost. These requirements provide the basis for the evaluation principles discussed in Section~\ref{sec:evaluation}.

\section{Datasets, Platforms, Baselines, and~Evaluation}
\label{sec:evaluation}

Active visual perception cannot be evaluated through a single
accuracy measure because the system changes the observations on
which later perception and task decisions depend. A~complete
evaluation must determine whether the selected action produces
better visual evidence, whether the new evidence improves the
downstream task, and~whether the improvement justifies the motion,
computation, communication, energy, and~safety costs required to
obtain it. A~method may achieve a more complete reconstruction while
following an impractically long camera path. Another method may
complete a task successfully because it uses a stronger perception
backbone rather than a better sensing policy. Final accuracy or task
success alone therefore cannot isolate the contribution of active
observation~selection.

The experimental requirements also differ across tasks. Active
reconstruction requires geometric ground truth and controllable
viewpoints. Active localization and exploration require an
environment in which the current movement changes future
observations. Search and tracking require targets whose visibility
changes with camera or platform actions. Manipulation requires
observation control to remain connected to reachability and physical
task execution. Collaborative perception additionally requires
controllable communication, relative pose error, delay, and~information overlap. The~following analysis distinguishes the
resources used to support these experiments from the protocols used
to judge the sensing~decision.
\newpage
\subsection{Representative Datasets and Interactive~Environments}

Active reconstruction studies commonly use synthetic meshes,
ShapeNet style object collections, Middlebury data, custom CAD
models, and~small collections of real objects
~\cite{dunn2009next,peralta2020next,chen2024gennbv,
dhami2023pred}. These resources provide accurate camera poses,
surface geometry, and~depth ground truth. They make it possible to
measure surface coverage, geometric error, reconstruction
completeness, and~the contribution of each additional view.
However, many experiments select from predefined camera positions
around an isolated object. Such settings test whether the policy can
identify geometrically useful views, but~they do not fully test
whether those views are reachable by a physical~robot.

Active localization and exploration use recorded trajectories,
custom simulation maps, indoor environments, outdoor routes, and~physical mobile robots
~\cite{ahmed2023active,placed2022explorb,
bonetto2022irotate,wang2023active}. KITTI, Freiburg, Intel Research
Lab, and~MIT \hl{Stata} 
 Center data can support evaluation of pose and map
estimation. Their trajectories are fixed, however, which means that
a policy cannot alter the route or acquire a different observation.
They can therefore evaluate a localization or mapping component, but~they cannot by themselves evaluate active route or viewpoint
selection. Interactive simulation or a physical platform is required
when the research question concerns where the robot should move,
look, or~revisit.

Search and navigation have a more developed set of shared
interactive environments. The~Active Vision Dataset provides dense
RGB and depth observations at connected positions in real indoor
scenes and supports controlled selection among neighbouring
viewpoints~\cite{ammirato2017dataset}. Habitat provides efficient
embodied simulation in three dimensional indoor environments
~\cite{savva2019habitat}, while RoboTHOR connects simulated scenes
with corresponding physical environments for transfer evaluation
~\cite{deitke2020robothor}. These environments allow different
policies to start from comparable states and change future
observations through movement and gaze actions. Their results still
depend on the available action set, object annotations, sensor model,
and perception~backbone.

Manipulation studies use tabletop simulation, custom object
collections, physical manipulators, and~general robot manipulation
environments. Recent work has also introduced resources designed
more directly for active observation. EFM10 evaluates exploratory
and focused manipulation, while \hl{ActiveViewPose 200K} 
 and
\hl{ActiveManip Bench} evaluate active viewpoint acquisition and
manipulation under changing visual evidence
~\cite{he2026towards,liu2026sapave}. These resources move beyond the
assumption that the robot begins from a favourable view. Their main
difficulty is that final task performance remains jointly affected
by perception, viewpoint selection, motion planning, and~physical
control.

Active tracking and collaborative perception often require custom
interactive environments because fixed videos and fixed multiview
datasets do not allow a policy to redirect a camera or reassign a
sensor. Tracking studies therefore use controllable pan, tilt, and~zoom cameras, camera networks, simulated moving targets, and~physical
video systems
~\cite{murray1994motion,cai2016persistent,
chen2024optimizing,casao2024distributed}. Collaborative driving
studies use resources such as V2X-Sim, OPV2V, V2XSet, and~DAIR-V2X
to examine complementary observations and communication efficient
fusion
~\cite{hu2022where2comm,yang2023how2comm,
huang2024actformer}. These resources provide multiview evidence, but~some of them offer only limited control over how the next
observations are generated. They are therefore more suitable for
evaluating active message or region selection than for evaluating
joint vehicle motion and observation control. Table~\ref{tab:datasets} summarizes representative datasets and interactive environments together with their support for active evaluation and main~limitations.

\begin{table}[H]
\caption{\hl{Representative} 
 datasets and interactive environments used
in active visual perception.}
\label{tab:datasets}

\begin{adjustwidth}{-\extralength}{0cm}
\centering
\renewcommand{\arraystretch}{1.18}
\setlength{\tabcolsep}{3.5pt}
\footnotesize

\begin{tabularx}{\fulllength}{@{}
>{\hsize=0.76\hsize\linewidth=\hsize
  \raggedright\arraybackslash}X
>{\hsize=1.26\hsize\linewidth=\hsize
  \raggedright\arraybackslash}X
>{\hsize=1.00\hsize\linewidth=\hsize
  \raggedright\arraybackslash}X
>{\hsize=0.98\hsize\linewidth=\hsize
  \raggedright\arraybackslash}X@{}}

\toprule
\textbf{Task} &
\textbf{Representative Resources} &
\textbf{Support for Active Evaluation} &
\textbf{Main Limitation} \\
\midrule

Active reconstruction
~\cite{dunn2009next,peralta2020next,
chen2024gennbv,dhami2023pred}
&
Synthetic meshes, ShapeNet style objects, Middlebury data,
custom CAD models, and~real object collections
&
Accurate geometry and controllable views support coverage,
completeness, and~view contribution analysis
&
Viewpoints are often predefined and ignore reachability,
collision, and~motion error
\\
\midrule

Active localization and exploration
~\cite{ahmed2023active,placed2022explorb,
bonetto2022irotate}
&
KITTI, Freiburg, Intel Research Lab, MIT Stata Center,
custom simulators, and~mobile robots
&
Interactive settings support route, revisit, coverage,
and loop closure decisions
&
Recorded trajectories cannot evaluate action selection,
while custom environments limit direct comparison
\\
\midrule

Search and navigation
~\cite{ammirato2017dataset,savva2019habitat,
deitke2020robothor}
&
Active Vision Dataset, Habitat, RoboTHOR,
and related indoor embodied environments
&
Policies can change position and gaze and acquire
task dependent observations
&
Results depend on annotations, action discretization,
sensor assumptions, and~the perception backbone
\\
\midrule

Active tracking
~\cite{cai2016persistent,chen2024optimizing,
casao2024distributed}
&
Controllable camera networks, pan tilt zoom systems,
simulated targets, and~physical video platforms
&
Camera orientation, assignment, and~target handover
can affect future visibility
&
Custom target dynamics and camera models reduce
comparability across studies
\\
\midrule

Robotic manipulation
~\cite{jauhri2024active,he2026towards,
liu2026sapave}
&
Tabletop simulation, physical manipulators, EFM10,
ActiveViewPose 200K, and~ActiveManip Bench
&
Viewpoint and interaction actions can be evaluated
together with task execution
&
Perception, sensing, planning, and~control gains
remain difficult to separate
\\
\midrule

Collaborative perception
~\cite{hu2022where2comm,yang2023how2comm,
huang2024actformer,bai2024multi}
&
V2X-Sim, OPV2V, V2XSet, DAIR-V2X,
multirobot maps, and~camera networks
&
Supports complementary observation, message selection,
and communication cost analysis
&
Accurate poses, synchronization, and~stable communication
can overestimate collaborative gains
\\

\bottomrule
\end{tabularx}
\end{adjustwidth}
\end{table}
\unskip

\subsection{Simulation and Physical~Platforms}

Simulation provides repeatable interaction because different policies
can begin from the same initial state while scene layout, target
position, occlusion, sensor noise, and~action budget remain
controlled. It also supports the large number of trials required by
reinforcement learning and foundation model-based policies.
Embodied environments are suitable for search and navigation,
robotics simulators support manipulation and exploration, and~aerial or game engine environments can support UAV and active camera
experiments. Their main advantage is not visual realism alone. Their
essential property is that the selected action changes the next
observation.

The action interface of the simulator strongly affects the result.
A small discrete action space limits the number of possible
viewpoints and makes search easier. Continuous movement introduces
reachability, trajectory length, and~control constraints. Simulated
depth, pose, and~visibility are often more reliable than physical
measurements. Some environments also execute camera rotation or
platform motion without realistic acceleration, blur, vibration, or~delay. A~policy may therefore appear efficient because the platform
can acquire views that would be slow or unstable in a physical
system.

Physical platforms reveal whether the selected observation can
actually be obtained. Mobile robots expose wheel slip, localization
drift, obstacle clearance, and~limited camera placement.
Manipulators introduce joint limits, self occlusion, collision, and~coupling between the sensing pose and the following task action.
UAVs introduce battery, wind, stability, and~safety constraints.
Camera networks introduce synchronization and calibration error,
while multirobot systems additionally require relative localization
and communication. These experiments provide stronger evidence of
feasibility, but~they are usually smaller and less repeatable than
simulation~studies.

Simulation and physical evaluation should therefore be interpreted
together. Simulation can compare policies under controlled
conditions, while physical experiments determine whether the
predicted observation value survives sensor and execution error.
A small physical demonstration cannot establish broad robustness,
but simulation alone cannot establish that an informative viewpoint
is reachable, timely, or~safe.

\subsection{Baselines and Evaluation~Metrics}

A passive or fixed observation baseline is required to determine
whether changing the sensing process is necessary. This baseline
keeps the viewpoint, route, sensor schedule, or~communication policy
unchanged. A~random baseline tests whether the proposed action
selection rule performs better than uninformed sampling. A~uniform
baseline distributes observations evenly and is especially relevant
to reconstruction, inspection, and~monitoring. A~greedy baseline
selects the action with the largest immediate value and helps
determine whether longer horizon reasoning provides an additional
benefit. When feasible, an~oracle or privileged information baseline
can estimate an upper bound, but~its access to ground truth must be
reported~clearly.

Metrics should be reported at several levels. Observation level
metrics measure what changed after the sensing action. Examples
include target visibility, valid depth, surface coverage, entropy,
pose covariance, detection confidence, and~occlusion reduction.
Task level metrics measure the effect on reconstruction, localization,
search, tracking, manipulation, or~collaboration. Examples include
geometric error, ATE, RPE, search success, target loss, MOTA, IDF1,
grasp success, and~detection accuracy. Action level metrics measure
the effort required to obtain the evidence, including view count,
path length, camera rotation, action count, and~completion time.
System level metrics measure computation, memory, energy,
communication volume, transmission delay, and~safety~violations.

These levels should not be replaced by a single weighted score.
Two policies may achieve the same task success while using different
travel distances or communication volumes. Two reconstruction
methods may use the same number of views while producing very
different camera paths. Reporting the underlying terms separately
allows readers to determine whether an apparent gain remains useful
under another platform or resource budget. Table~\ref{tab:evaluation} summarizes the main baselines and measurements required for closed-loop~evaluation.

\begin{table}[H]
\caption{\hl{Baselines} 
 and measurements required for closed loop
evaluation of active visual perception.}
\label{tab:evaluation}

\begin{adjustwidth}{-\extralength}{0cm}
\centering
\renewcommand{\arraystretch}{1.18}
\setlength{\tabcolsep}{3.5pt}
\footnotesize

\begin{tabularx}{\fulllength}{@{}
>{\hsize=0.86\hsize\linewidth=\hsize
  \raggedright\arraybackslash}X
>{\hsize=1.10\hsize\linewidth=\hsize
  \raggedright\arraybackslash}X
>{\hsize=1.16\hsize\linewidth=\hsize
  \raggedright\arraybackslash}X
>{\hsize=0.88\hsize\linewidth=\hsize
  \raggedright\arraybackslash}X@{}}

\toprule
\textbf{Evaluation Question} &
\textbf{Required Baseline or Measurement} &
\textbf{What It Reveals} &
\textbf{Common Confounding Factor} \\
\midrule

Is active sensing necessary?
&
Passive or fixed observation baseline
&
Whether changing the observation improves perception
or the downstream task
&
The active method may also use a stronger perception model
\\
\midrule

Is the selection rule informative?
&
Random and uniform action baselines
&
Whether task or uncertainty information improves action choice
&
Different candidate action sets may favour one method
\\
\midrule

Does planning horizon matter?
&
Greedy or immediate utility baseline
&
Whether reasoning about later observations improves the result
&
Longer planning may use more computation or privileged prediction
\\
\midrule

Did the observation itself improve?
&
Visibility, coverage, uncertainty, valid depth,
or pose quality before and after the action
&
The direct contribution of the acquired evidence
&
The metric may be poorly calibrated or unrelated to the task
\\
\midrule

Did the task improve?
&
Reconstruction, localization, search, tracking,
manipulation, or~detection performance
&
Whether the new evidence changes the final task outcome
&
Control quality and the perception backbone also affect the result
\\
\midrule

Was the improvement efficient?
&
View count, path length, action count, time,
energy, computation, and~communication
&
The resources required to obtain the reported improvement
&
A combined score can hide which resource caused the difference
\\
\midrule

Is the policy robust?
&
Sensor noise, pose error, actuation delay, packet loss,
dynamic occlusion, and~unseen scenes
&
Whether the sensing decision remains useful outside ideal settings
&
Perturbation ranges and failure definitions differ across studies
\\
\midrule

Did the system stop appropriately?
&
Marginal gain, additional action benefit,
and stopping error
&
Whether further sensing would change the task decision
&
Many studies use a fixed action budget instead of a learned stop rule
\\

\bottomrule
\end{tabularx}
\end{adjustwidth}
\end{table}
\unskip

\subsection{Requirements for Closed-Loop~Evaluation}

A fair comparison should keep the perception backbone, initial
observation, candidate action space, sensing budget, and~task
controller consistent whenever the research question concerns the
sensing policy. Otherwise, a~higher final result may arise from a
better detector, depth estimator, semantic model, or~controller.
When identical components cannot be used, the~differences should be
reported and supported through module level~ablations.

Evaluation should also measure the state immediately before and
after the sensing action. This comparison reveals whether the
selected observation actually reduced occlusion, uncertainty, pose
error, or~another information deficit. Final task success remains
important, but~it should be reported together with this intermediate
change. Without~the intermediate measurement, the~experiment cannot
show whether the task improved because of better visual evidence or
because another part of the system compensated for a weak sensing
decision.

Privileged information must be identified explicitly. Ground truth
maps, exact camera poses, perfect depth, known target locations, and~complete communication graphs can be useful for isolating a planning
component, but~they are not available to a deployed system.
Experiments should distinguish results obtained with estimated state
from results obtained with ground truth state. Robustness tests
should then introduce pose error, sensor noise, actuation delay,
dynamic occlusion, packet loss, and~unfamiliar environments
according to the deployment~setting.

Finally, the~experiment should define when the system has acquired
enough evidence. A~fixed number of views makes comparison easier,
but it does not show whether the policy can recognize that another
observation is unnecessary. A~stopping policy should be evaluated
through the benefit of the final additional action, the~cost of
unnecessary observations, and~the frequency with which the system
stops before acquiring evidence needed by the task. This decision is
part of active visual perception because sensing remains useful only
while a new observation can change the subsequent estimate or
action.

Existing resources demonstrate that active observation can improve
reconstruction, localization, search, tracking, manipulation, and~collaborative perception. However, no single dataset or platform
currently provides common tasks, sensing actions, physical
constraints, and~resource measurements across all of these settings.
The strongest evidence therefore comes from evaluations that combine
shared interactive environments, controlled component comparisons,
physical feasibility tests, and~separate reporting of observation
benefit, task improvement, and~acquisition~cost.

\subsection{Applicability Boundaries of Datasets and~Platforms}

{The applicability of each dataset or platform is determined not only by its scene domain and annotations, but~also by whether it permits the sensing intervention claimed by the evaluated method. Synthetic meshes, CAD object collections, and~densely sampled multiview datasets are suitable for evaluating geometric reconstruction, coverage, and~next-best-view quality because they provide accurate geometry and controllable camera poses. Their applicability is limited when the research question involves reachability, collision avoidance, motion error, material-dependent sensing, or~other physical constraints. Recorded trajectory datasets are suitable for evaluating localization, mapping, detection, and~perception under realistic sensor observations. However, because~their trajectories and observations are fixed, they cannot independently evaluate whether an active policy selects a better route, viewpoint, or~sensing action.}

{Interactive simulators are suitable for closed-loop search, navigation, exploration, reinforcement learning, and~long-horizon planning because an action can change the next observation under repeatable conditions. Their conclusions remain bounded by the fidelity of the action space, sensor model, dynamics, communication model, and~visual domain. Real-world multiview, tracking, manipulation, and~collaborative datasets provide realistic appearance, noise, occlusion, and~communication conditions, but~fixed camera layouts, predefined trajectories, or~recorded messages may restrict evaluation to observation fusion, message selection, or~perception rather than active motion control. Physical robot and camera platforms are required when the research claim concerns reachability, control delay, calibration, energy consumption, physical interaction, or~safety. Nevertheless, physical experiments are usually platform-specific, costly, and~less repeatable, and~therefore cannot alone establish broad generalization. Dataset selection should consequently match the level of the research claim, and~strong evidence should combine controlled simulation, realistic recorded data, and~physical-platform validation whenever deployment feasibility is claimed.}

\subsection{Practical Checklist for Closed-Loop~Evaluation}

{The preceding analysis can be consolidated into the following reusable checklist for designing and reporting closed-loop active visual perception experiments.}

\begin{itemize}

\item \textbf{\hl{Closed-loop interaction.}} 
The selected sensing action should actually change the subsequent observation, and~the acquired evidence should update the state used for the next~decision.

\item \textbf{\hl{Appropriate evaluation setting.} }
The dataset, simulator, or~physical platform should permit the sensing intervention claimed by the method and should match the intended level of the research~claim.

\item \textbf{\hl{Controlled comparison.} }
The initial observation, perception backbone, candidate action space, sensing budget, task controller, and~stopping condition should be kept consistent whenever possible. Any unavoidable differences should be reported~explicitly.

\item \textbf{\hl{Baselines and ablations.} }
The method should be compared with appropriate passive or fixed, random, uniform, greedy, and, when feasible, oracle baselines. Component-level ablations should isolate the contributions of state estimation, observation-value prediction, action selection, and~control.

\item \textbf{\hl{Observation and task improvement.} }
Evaluation should report both the change in visual evidence before and after the sensing action and the resulting effect on the downstream task, rather than relying only on the final task~score.

\item \textbf{\hl{Cost, feasibility, and~robustness.} }
View count, motion distance, completion time, computation, energy, communication, reachability, collision risk, and~execution failures should be reported separately. Relevant sensor, state-estimation, execution, environmental, and~communication disturbances should also be~tested.

\item \textbf{\hl{Information access.} }
Ground-truth states, exact poses, perfect depth, privileged maps, complete communication graphs, and~any other information unavailable during deployment should be disclosed~explicitly.

\item \textbf{\hl{Stopping and reproducibility.} }
Both premature stopping and unnecessary additional observations should be measured. The~number of trials, random seeds, failure criteria, system configurations, and~sufficient action--observation traces should be reported to support~reproduction.

\end{itemize}

\section{Representative~Applications}
\label{sec:applications}

\begin{figure}[H]

\includegraphics[width=0.86\textwidth]{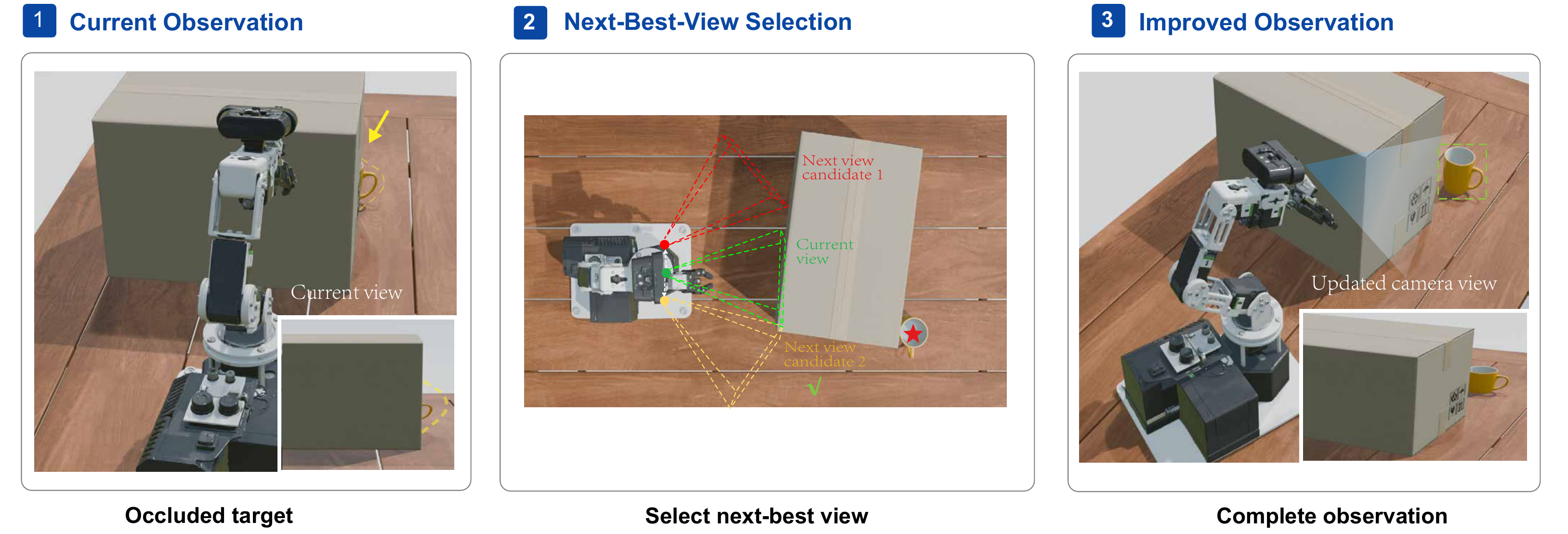}

\vspace{1mm}
{\small \hspace{60mm}(\textbf{a})}
\vspace{3mm}

\includegraphics[width=0.86\textwidth]{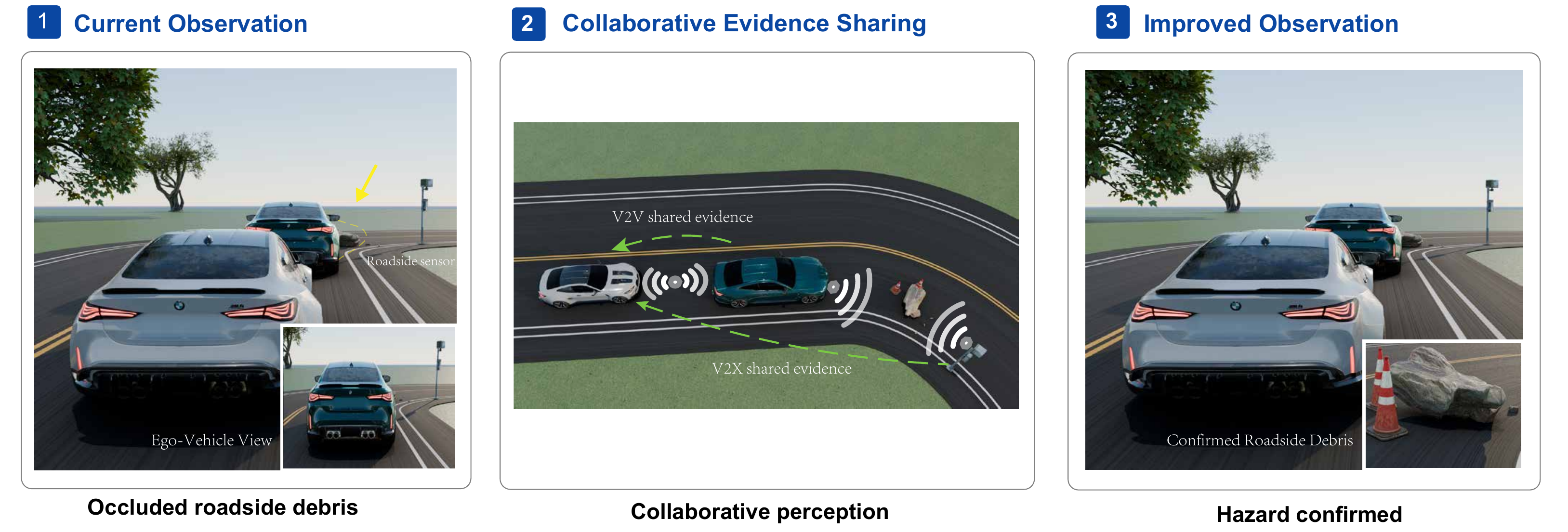}

\vspace{1mm}
{\small \hspace{60mm}(\textbf{b})}
\vspace{3mm}

\includegraphics[width=0.86\textwidth]{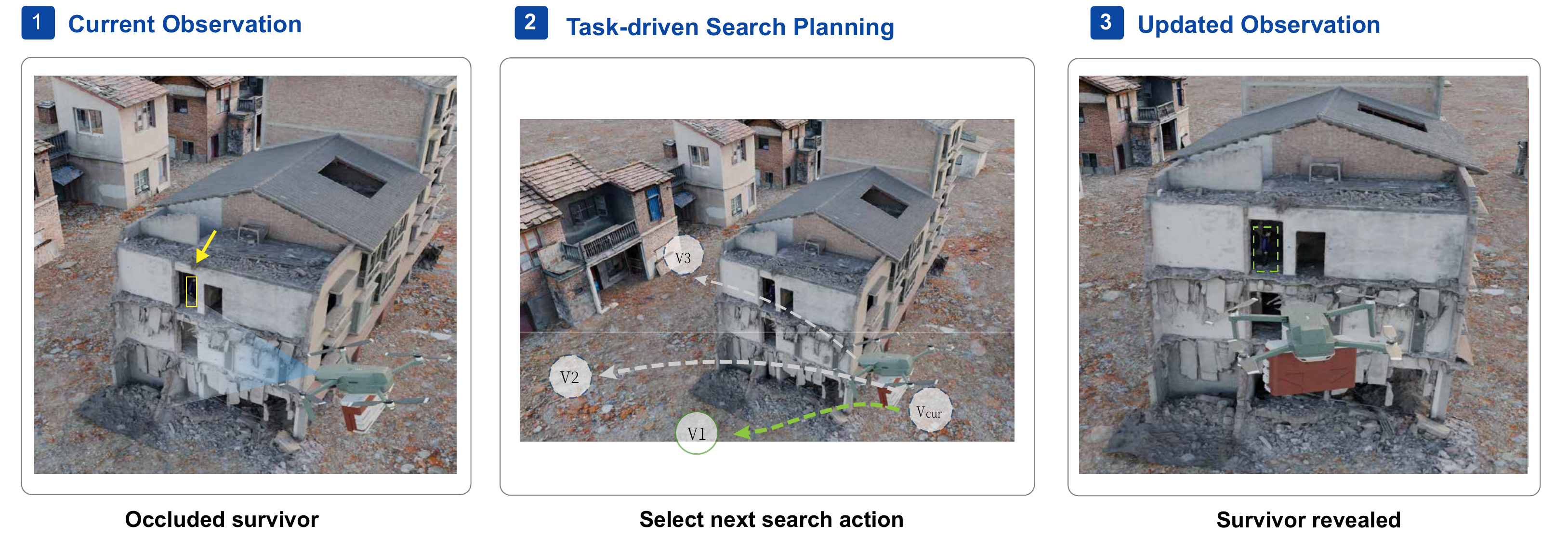}

\vspace{1mm}
{\small~\hspace{60mm}(\textbf{c})}

\caption{\hl{Representative active perception processes in robotic manipulation, collaborative driving, and UAV-assisted disaster rescue. In all subfigures, the yellow arrows point to the target, while the boxes in the lower-right corner of panels (a) and (b) show enlarged local views of the region observed from the current viewpoint. (a) Next-best-view selection for robotic manipulation, where the current and candidate viewpoints are distinguished by viewing cones with different colors; the asterisk marks the target, and the check mark indicates the selected viewpoint. (b) Collaborative driving, where the green dashed arrows indicate the sharing of complementary perceptual information between vehicles and between vehicles and the roadside sensor. (c) UAV-assisted disaster rescue, where the dashed arrows schematically illustrate three candidate search paths for acquiring task-relevant visual observations; the green dashed path indicates the selected search path.} 
}
\label{fig:application_scenarios}

\end{figure}

\hl{The role of active visual perception changes with the task and the environment in which the system operates, as summarized in} Table~\ref{tab:applications}. Robotic manipulation uses additional viewpoints to reveal structures required for physical interaction. Autonomous vehicles and mobile robots seek observations that improve safety, localization, and~decision-making in dynamic scenes. UAV systems adjust their flight and sensing strategies according to the urgency and spatial distribution of the task. Figure~\ref{fig:application_scenarios} illustrates three representative forms of active visual perception. These forms include next best view selection in robotic manipulation, complementary observation sharing in collaborative driving, and~task-guided perception in UAV assisted disaster~rescue.

\unskip

\subsection{Robotic~Manipulation}

In practical robotic manipulation, detecting an object does not necessarily provide the visual evidence required to act on it. In~cluttered picking, the~target may be recognizable while its graspable surfaces remain occluded, leaving the robot without a reliable collision-free grasp~\cite{WOS:001189842600006}. Similar difficulties arise when manipulation depends on an accurate object pose rather than object identity alone~\cite{an2024rgbmanip}. Transparent and specular objects further complicate deployment because their geometry may be poorly measured by depth sensors even when they are visible in an image~\cite{WOS:001676224400011}. The~information requirement also changes during execution, as~movement of the gripper or deformation of the object can alter what remains observable. Additional sensing must therefore be coordinated with the operation rather than treated as an independent preliminary stage. This creates an engineering tradeoff because moving the camera consumes time and workspace, while manipulating an occluder can change the scene that the robot must subsequently~handle.

\begin{table}[H]
\caption{\hl{Perception--action} 
 characteristics of representative application~settings.}
\label{tab:applications}
\begin{adjustwidth}{-\extralength}{0cm}
\centering
\renewcommand{\arraystretch}{1.16}
\setlength{\tabcolsep}{3pt}
\footnotesize
\begin{tabularx}{\fulllength}{@{}
>{\hsize=0.76\hsize\linewidth=\hsize\raggedright\arraybackslash}X
>{\hsize=1.02\hsize\linewidth=\hsize\raggedright\arraybackslash}X
>{\hsize=1.02\hsize\linewidth=\hsize\raggedright\arraybackslash}X
>{\hsize=1.10\hsize\linewidth=\hsize\raggedright\arraybackslash}X
>{\hsize=1.10\hsize\linewidth=\hsize\raggedright\arraybackslash}X@{}}
\toprule
\textbf{Application} & \textbf{Information Deficit} & \textbf{Controlled Action} & \textbf{Expected Evidence} & \textbf{Dominant Constraints} \\
\midrule
Robotic manipulation & Hidden grasp region, uncertain pose, or~unknown object state & Wrist or head motion, base motion, active-view arm, or~object interaction & Clearer affordance, improved pose, or~interaction-generated state evidence & Collision, reachability, irreversible action, latency, and~task~coupling \\
\midrule
Autonomous driving & Occluded road user, weak visibility, or~incomplete local sensing & Camera orientation, adaptive resolution, modality selection, or~collaborative query & Target confirmation, complementary geometry, or~remote hazard evidence & Real-time response, safety, weather, synchronization, and~pose~error \\\midrule
Human-centered systems & Occluded body, uncertain gesture, or~missing shared-task evidence & Gaze, camera orientation, robot repositioning, or~selective resolution & Clearer pose, attention, intention, or~shared-object state & Personal space, privacy, interpretability, consent, and~social~acceptance \\\midrule
Surveillance and security & Blind spot, target exit, identity ambiguity, or~limited resolution & Pan, tilt, zoom, camera assignment, or~target handover & Centered target, identity-preserving sequence, or~complementary views & Target competition, control delay, privacy, and~operator~accountability \\\midrule
UAV search and rescue & Unknown layout, hidden victim, or~inaccessible region & Flight path, altitude, orientation, revisit, or~team allocation & Victim evidence, damage map, route information, or~wide-area coverage & Battery, collision, urgency, wind, and~communication \\\midrule
Environmental monitoring & Sparse anomaly, changing conditions, or~uneven target distribution & Route, viewpoint, sensor schedule, or~spatially varying sampling & Focused anomaly evidence and long-term change observations & Energy, weather, revisit fairness, and~long-term~autonomy \\
\bottomrule
\end{tabularx}
\end{adjustwidth}
\end{table}

Existing methods address these difficulties by progressively expanding how the robot acquires missing evidence. Active vision systems adjust camera viewpoints to support manipulation, while visual servoing can guide observation toward less occluded regions~\cite{chuang2025active,lehnert20193d}. Active pose estimation connects additional observations to object manipulation, and~mobile manipulation extends this coordination to motion of the robot as a whole~\cite{an2024rgbmanip,jauhri2024active}. Probabilistic active grasping further links viewpoint selection to reconstruction uncertainty and online estimates of grasp success~\cite{WOS:001189842600006}. A~concrete example of accounting for sensing cost is HEAPGrasp, which uses an RGB camera attached to the manipulator to reconstruct object geometry from silhouettes and balances measurement accuracy against camera travel. Its real robot experiments demonstrate grasping across different optical properties while reducing camera travel and handling time relative to its counterpart using a predefined trajectory around the scene. This illustrates how observation planning can improve acquisition efficiency rather than simply collect more views~\cite{WOS:001676224400011}. When camera motion alone is insufficient, interactive methods manipulate objects to improve detection, expose unobserved surfaces, or~reveal hidden scene structure~\cite{le2008active,krainin2011autonomous,jiang2024roboexp}. The~Object-Aware Interactive Perception System implements selection between sensor repositioning and object manipulation on a \hl{dual-arm Baxter robot (Rethink Robotics, Boston, MA, USA) }equipped with RGB and depth cameras~\cite{WOS:001221645700001}. Weng~et~al. extend coordinated observation and interaction to a deformable bag using two \hl{UR10e arms (Universal Robots A/S, Odense, Denmark)}, with~one manipulating the bag and the other carrying a \hl{RealSense D435 camera (Intel Corporation, Santa Clara, CA, USA)}~\cite{weng2024interactive}. Recent vision language action methods further incorporate observation adjustment into task execution~\cite{liu2026activevla,liu2026sapave}.

{Nevertheless, these methods remain constrained by the relationship between informative observations and executable manipulation. A~useful camera pose must be reachable without compromising the subsequent operation, which makes coordinated perception and motion generation important in confined workspaces~\cite{jauhri2024active}. Avoiding unreliable depth sensing also does not eliminate perceptual assumptions, since HEAPGrasp still requires usable silhouette segmentation and camera pose information~\cite{WOS:001676224400011}. Interactive sensing adds another dependency because the action that reveals information changes the object state. In~the bag study, the~physical demonstration uses a restricted camera workspace and a visually marked opening, while separate simulation experiments report failures associated with excessive gripper motion and slower camera movement~\cite{weng2024interactive}. These conditions limit what the demonstrations establish about unrestricted deployment. Integrating observation and action within a learned policy does not by itself resolve such physical constraints. Engineering evaluation should therefore measure the complete sensing and manipulation cycle, distinguish improvements in visual evidence from changes in the perception or control model, and~test recovery when an observation or interaction fails. The~practical objective is to acquire enough evidence for reliable execution without allowing additional sensing to introduce disproportionate motion, delay, or~scene disturbance.}

\subsection{Autonomous Driving and Mobile~Navigation}

{Autonomous vehicles and mobile robots must make motion decisions before every relevant part of a changing environment can be observed. At~an occluded intersection, onboard cameras and range sensors cannot determine whether a hidden region is empty or contains a pedestrian, cyclist, or~vehicle about to enter the planned path. A~planner that treats every hidden region as occupied may stop unnecessarily or become trapped, whereas a planner that ignores unobserved road users accepts an unmodelled collision risk. Visual navigation has a related evidence problem. Feature poor walls, doorway transitions, moving objects, illumination changes, and~adverse weather can reduce stable visual correspondences and interrupt pose estimation even when the geometric route remains traversable~\cite{WOS:001173317800107,WOS:001531774200019,WOS:001542495900011}. The~operational question is therefore whether the current observation supports the next motion decision with sufficient spatial accuracy and~freshness.

Existing systems acquire complementary evidence through communication or controlled viewpoint changes. Learning-based collaborative perception methods such as Where2comm and How2comm select informative regions or communication partners instead of transmitting every observation~\cite{hu2022where2comm,yang2023how2comm}. In~a simulation study of occluded driving, a~V2X aware POMDP planner identifies hidden regions, represents possible unseen road users, selects external detections according to coverage, delay, and~sensor reliability, and~uses the selected evidence to update its belief before choosing a driving policy~\cite{WOS:001173317800107}. Engineering prototypes demonstrate parts of this process on physical systems. \hl{At the CETRAN autonomous vehicle test track, a roadside LiDAR system was connected to the Baidu Apollo autonomous driving platform (Baidu, Beijing, China) and Cohda Wireless communication hardware (Cohda Wireless Pty Ltd., Adelaide, Australia) under the SAE J2735 protocol.} 
 The~experimental vehicle consequently received detections of vehicles and pedestrians before they entered the field of view of its onboard camera~\cite{WOS:001160501800001}. A~more recent intelligent roadside unit integrates camera and LiDAR feature extraction on an FPGA with dual neural processing units and cellular V2X communication. Near~sensor computation and selective transmission reduced end to end perception latency in real roadside trials and simulations~\cite{WOS:001556085600036}. Mobile platforms can instead change the observation itself. Semantic aware navigation for UAVs and outdoor active view planning for visual SLAM select routes or viewpoints that expose useful scene information~\cite{bartolomei2021semantic,wang2023active}, while adaptive camera view adjustment-based on the distribution of visual features has improved localization robustness in simulated and real environments~\cite{WOS:001531774200019}.

These results do not yet establish a complete deployment chain from external observation to safety certified vehicle control. The~CETRAN experiments were conducted on an enclosed test track, and~the roadside detections were visualized and compared with onboard images rather than being used directly by the planning and control modules~\cite{WOS:001160501800001}. The~\hl{V2X aware POMDP} 
was evaluated in a proprietary simulator, while an asynchronous collaborative detection framework was evaluated on the V2X-Sim dataset rather than on communicating physical vehicles~\cite{WOS:001173317800107,WOS:001189499000001}. The~latter study nevertheless exposes important deployment assumptions because exchanged detections still require a common time reference and accurate transformations between agents. Buildings can reduce packet delivery, communication delay can make moving object observations stale, and~pose or calibration errors can place shared objects in the wrong part of the scene. Authors of active view planning research further report failures when learned policies encounter difficult transitions and insufficient visual texture, together with a need for human supervised training~\cite{WOS:001542495900011}. At~the system level, camera actuation and viewpoint changing motion also consume time and energy and may conflict with collision avoidance or traffic constraints. Active perception is therefore useful only when expected information gain is balanced against communication, computation, calibration, and~motion costs, and~when the resulting observation remains valid when the downstream decision is executed.}

\subsection{Human Centered Interaction and Active Surveillance}

{Human centered robots and active surveillance systems must obtain task-relevant evidence about body pose, gaze, gesture, identity, and~shared objects while people continue to move. Fixed cameras can lose hands, faces, or~body parts because of self occlusion, restricted fields of view, and~backlighting. In~collaborative work, such missing evidence prevents a robot from reliably determining whether a person has entered its motion space or issued a gesture command. In~surveillance, it can interrupt a trajectory and leave the identity or attributes of a target unresolved. The~engineering problem is therefore not simply to detect a person, but~to maintain sufficiently complete and timely observations for collision avoidance, interaction decisions, and~target management~\cite{WOS:001293884400001,casao2024distributed}.

Earlier studies established active visual search, gaze-based interfaces, and~gesture interaction~\cite{halverson2011computational,reale2011multi}, while visual attention and proactive scene understanding allowed robots to redirect sensing as an interaction developed~\cite{das2015supporting,ognibene2022active,fan2022vision}. Recent engineering systems connect these principles more explicitly to physical sensing and camera motion. A~multi-camera system dynamically captures RGB-D observations from important regions around a robotic arm and fuses partially visible body parts, which improves human sensing and supports obstacle avoidance under localized occlusion~\cite{WOS:001293884400001}. For~active surveillance, a~robotic camera array adjusts the positions and orientations of multiple cameras through cooperative reinforcement learning and has been evaluated for face detection and recognition in both simulation and real experiments~\cite{WOS:001285340000073}. Hybrid networks can also use fixed cameras for global person tracking and robot mounted cameras for close views that reveal task-specific attributes, although~the reported evaluation of this arrangement remains photorealistic simulation~\cite{casao2024distributed}. \hl{These implementations extend earlier PTZ tracking and camera allocation methods~\cite{foresti2005active,cai2016persistent,yang2026towards,mo2026unitrack}.} 
A~recent PTZ surveillance system further integrates image acquisition, object detection, and~feedback control in a single processing loop and empirically examines how delayed observations and detector errors affect camera control~\cite{WOS:001500451600001}.

These systems nevertheless remain constrained by the difference between an informative observation and an executable or acceptable one. Multi-view RGB-D sensing requires spatial calibration, synchronized observations, reliable association of body parts, and~sufficient computation for fusion. Moving cameras additionally introduce actuator speed, mechanical rotation, communication delay, and~control stability constraints. The~PTZ study directly reports the effects of network latency, detection accuracy, and~controller behaviour, while the hybrid mobile camera study is limited to simulation and should not be interpreted as a field deployment~\cite{WOS:001500451600001,casao2024distributed}. Privacy introduces another constraint because following a person or acquiring a closer view can capture information unrelated to the current task. A~privacy-oriented alternative uses a 60 GHz industrial FMCW radar to estimate body pose and recognize gesture commands without relying on ordinary video images, but~its evidence comes from a controlled dataset rather than closed-loop deployment in an operating production system~\cite{WOS:001542416000050}. A~separate study using the \hl{Temi mobile robot (Robotemi Ltd., Tel Aviv, Israel)} and the \hl{Ameca humanoid robot (Engineered Arts Ltd., Falmouth, United Kingdom)} in a controlled care scenario shows that visible indications of sensor deactivation and the ability to move a robot away influence perceived control and trust~\cite{WOS:001290125000001}. At~the system level, these findings imply that active observation policies must jointly consider perceptual benefit, calibration error, observation age, motion and computation costs, competition among targets, and~the user’s ability to understand or restrict sensing.}

\subsection{UAV Search, Rescue, and~Reconnaissance}

{UAV search, rescue, and~reconnaissance systems must acquire observations that are geographically broad enough to locate relevant regions and sufficiently detailed to support an operational decision. High altitude images cover large areas but make people and other critical objects appear small against cluttered backgrounds, while lower flight provides more detail at the cost of a narrower footprint and greater exposure to obstacles. Forest canopies, damaged buildings, smoke, and~poor illumination can prevent a top down camera from observing the ground, and~the absence of reliable positioning makes an otherwise valid detection difficult to associate with a rescue location. Missing or poorly localized evidence can therefore prevent the system from confirming a possible victim, prioritizing an access route, or~determining whether an area should be revisited~\cite{WOS:001558585600001,WOS:001638088100005,WOS:001160484400001}.}

{Earlier UAV methods established semantic guidance, collaborative search, formation control, and~multi-robot perception as mechanisms for directing sensing rather than applying uniform coverage alone~\cite{bartolomei2021semantic,queralta2020collaborative,tallamraju2019active,zhou2022multi}. Recent systems connect these ideas to more explicit engineering configurations. For~forest search, a~next best view planner evaluates how candidate camera positions and orientations expose surfaces that remain hidden by vegetation, and~real environment experiments show that the selected oblique views improve observation beneath the canopy~\cite{WOS:001638088100005}. A~heterogeneous rescue prototype combines UAVs for wide area observation with ground robots for closer inspection. RGB and thermal images are processed by FPGA-based edge hardware using a \hl{YOLO detector}, 
allowing suspected survivors to be presented to rescuers without sending every image to a centralized computing platform~\cite{WOS:001283693900001}. Multi UAV systems can also use onboard cameras to estimate neighbouring vehicles, construct a time varying position belief field, and~change their headings to reduce uncertainty or recover targets that have left the limited field of view. This architecture has been evaluated in both simulation and real flight experiments under communication denial~\cite{WOS:001508118600048}. For~persistent reconnaissance, multiple quadrotors can lower their altitude above high priority regions to obtain higher resolution observations while maintaining wider coverage elsewhere, with~distributed waypoint generation connecting changes in regional importance to subsequent flight decisions~\cite{WOS:001168273300001}.

The available evidence nevertheless remains narrower than a mature emergency deployment. The~forest next best view study includes real environment experiments with hidden objects, but~it does not report an operational search for actual disaster victims. The~heterogeneous UAV and ground robot system is a proof of concept whose components communicate through a shared wireless network, while the communication denied swarm experiments primarily demonstrate navigation, formation, and~target recovery rather than a complete rescue mission~\cite{WOS:001638088100005,WOS:001283693900001,WOS:001508118600048}. A~separate framework uses thermal sensing, LiDAR inertial localization, and~belief-based planning to navigate under degraded visibility and unavailable satellite positioning, but~its physical validation is a controlled indoor rescue style task rather than an outdoor disaster deployment~\cite{WOS:001160484400001}. Small person detection has also been demonstrated on embedded hardware and tested across different flight altitudes, yet this evidence comes mainly from image datasets and detector evaluation rather than a closed-loop autonomous search~\cite{WOS:001558585600001}. At~the system level, these validation conditions imply that camera resolution, localization drift, calibration, communication loss, actuator response, wind, payload, battery capacity, and~onboard computation can all change whether an informative viewpoint is reachable before its information becomes obsolete. Lowering altitude or revisiting an uncertain region may improve local evidence, but~it consumes flight time and reduces the coverage available to the rest of the mission. Active observation policies for rescue and reconnaissance must therefore evaluate information gain together with flight feasibility, observation age, communication reliability, and~the cost of leaving other regions unobserved.}

\subsection{Environmental~Monitoring}

{Environmental monitoring must convert sparse, partially observed, and~time varying signals into decisions about where to revisit, where to collect a physical sample, and~when to report a meaningful change. In~greenhouse crops, leaves and fruit clusters can hide the plant nodes required for harvesting and maintenance; so, a single image may not provide enough geometric evidence for manipulation. In~lakes and reservoirs, measurements collected only at the surface or along a fixed route may miss localized water quality changes and their variation with depth. Repeated underwater surveys introduce a further difficulty because the system must distinguish an actual environmental change from differences caused by viewpoint, localization drift, illumination, or~detection error. Fixed routes and uniform sampling therefore provide predictable coverage but cannot guarantee the target visibility, spatial confirmation, or~temporal correspondence required for subsequent intervention~\cite{wang2025multiple,WOS:001805473800032,WOS:001730242900001}.

Existing systems address these information deficits by using provisional observations to determine the next sensing action. Multimodal science mission planners can select both a motion path and a sensing modality according to their expected contribution under a limited mission budget~\cite{arora2019multi}. Simulated plant inspection has similarly shown how local next best view planning can direct the camera toward occluded plant nodes~\cite{burusa2024gradient}. This principle has subsequently been tested in an unmodified tomato greenhouse, where an RGBD camera collects observations from planned viewpoints, multiple object tracking associates plant nodes across views, and~the resulting three dimensional semantic representation guides the next camera pose. The~additional views reveal nodes hidden in earlier images and improve detection and localization relative to predefined viewpoints~\cite{wang2025multiple}. In~aquatic monitoring, an~Internet of Drones system combines water quality and depth sensors with wireless telemetry and a physical multisampler. Measurements of dissolved solids, depth, and~estimated chlorophyll conditions can trigger additional sampling, while the accumulated observations support spatial mapping and depth profiling during field tests at a Lake Erie test site~\cite{WOS:001805473800032}. At~a longer temporal scale, an~autonomous underwater inspection framework uses semantic observations to refine its coverage path and compares maps from different sessions to revisit and localize possible changes. Its buoy monitoring experiments demonstrate how new observations can influence both the current route and the interpretation of later surveys~\cite{WOS:001730242900001}. Semantic route selection has also been studied for aerial robots in photorealistic simulation, where task-relevant objects rather than travel distance alone determine the preferred observation regions~\cite{bartolomei2021semantic}.

These systems also expose assumptions that become fragile on physical platforms. Information gain and semantic priority are useful only when the selected viewpoint is reachable, the~sensor remains calibrated, and~the new measurement can be registered consistently with earlier observations. Greenhouse systems must tolerate occlusion, natural illumination changes, sensor noise, camera pose uncertainty, and~the motion time required to obtain supplementary views. Water monitoring must additionally account for sensor drift, delayed responses, depth dependent variation, communication coverage, payload capacity, and~the energy consumed when an anomaly triggers further sampling. Multi robot source localization demonstrates that switching from broad coverage to concentrated active sensing can accelerate estimation, but~its reported hardware in the loop and real time experiments use a radio source to emulate a gas field and therefore do not yet establish performance in a real pollutant plume~\cite{tran2025active}. Reactive benthic monitoring further shows that false detections can repeatedly divert an underactuated vehicle from area coverage. The~proposed recovery mechanism was evaluated in simulation, while the associated field evidence comes from a reanalysis of earlier trials whose controller experienced false positives and navigation failures~\cite{WOS:001830764600001}. Environmental active perception must therefore evaluate observation value together with calibration reliability, registration error, onboard computation, communication, and~the motion and energy costs of revisiting uncertain regions.}

\section{Key Challenges in Active Visual~Perception}
\label{sec:challenges}

\begin{figure}[H]
    \includegraphics[width=\linewidth]{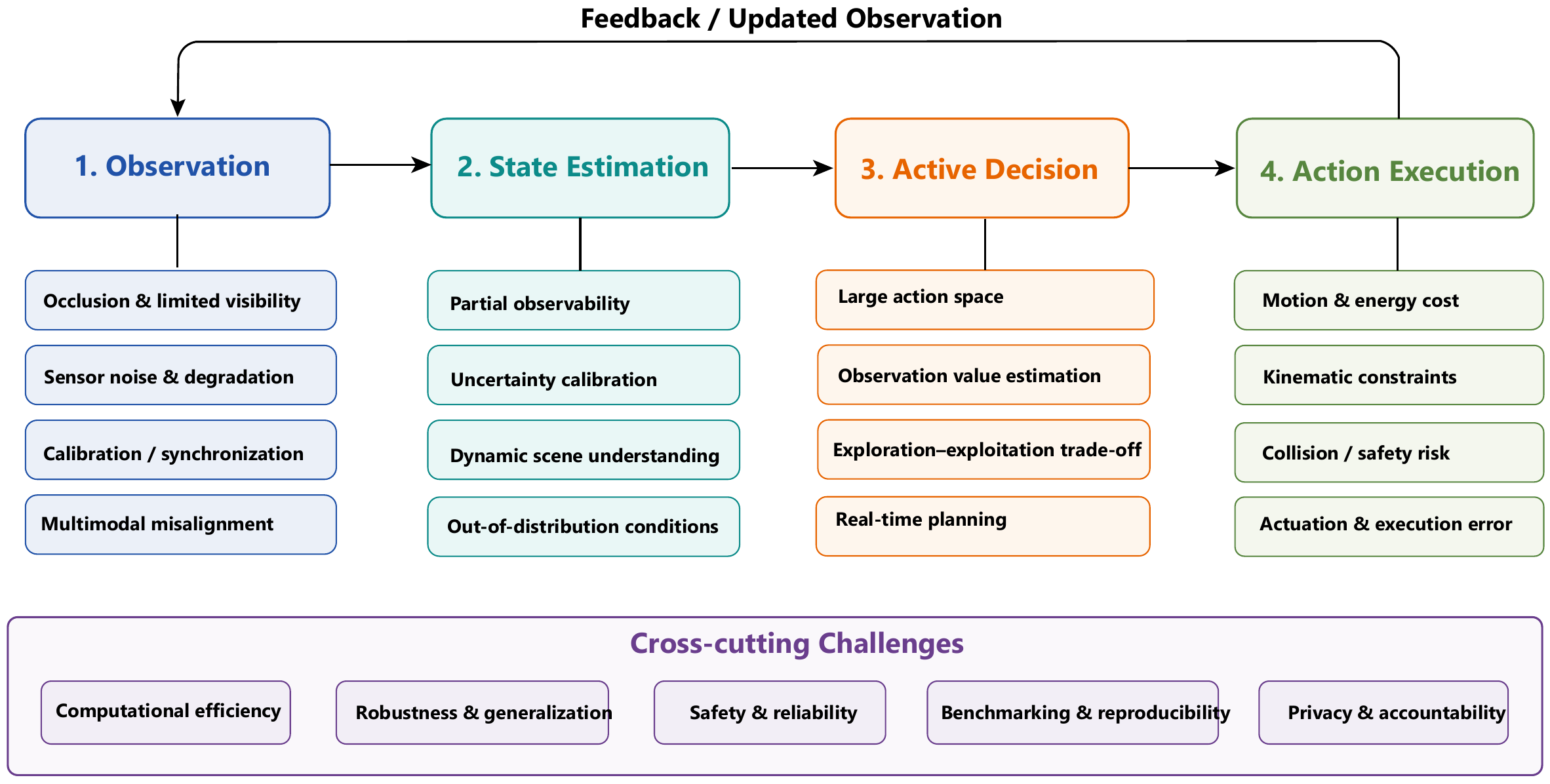}
    \caption{\hl{Interacting} 
 challenges in the active visual perception~loop.}
    \label{fig:challenges}
\end{figure}

Figure~\ref{fig:challenges} summarizes how these challenges interact within the active visual perception loop, while Table~\ref{tab:challenges} compares their causes, consequences, and~unresolved gaps. 

Active visual perception allows a system to improve its current information by changing the viewpoint, sensor configuration, or~source of observation. This capability also gives perception errors more complex consequences. In~passive vision, an~incorrect estimate may affect only the current output. An~active system can use the same error to select its next observation, causing later data to move further away from the evidence that is actually needed. View prediction, action execution, and~state update are mutually dependent, and~an error in one stage changes the visual evidence available to the stages that~follow.

Existing methods can select informative viewpoints, plan exploration routes, and~coordinate multiple sensors in controlled settings. Physical systems must nevertheless operate with incomplete observations, changing environments, motion constraints, limited computation, and~sensor error at the same time. The~central difficulty is no longer limited to finding an observation with high predicted value. The~complete perception and action process must remain reliable under these~conditions.

\begin{table}[H]
\caption{\hl{Main} 
 challenges and unresolved gaps in active visual~perception.}
\label{tab:challenges}
\begin{adjustwidth}{-\extralength}{0cm}
\centering
\renewcommand{\arraystretch}{1.18}
\setlength{\tabcolsep}{3pt}
\footnotesize

\begin{tabularx}{\fulllength}{@{}
>{\hsize=0.78\hsize\linewidth=\hsize
  \raggedright\arraybackslash}X
>{\hsize=1.05\hsize\linewidth=\hsize
  \raggedright\arraybackslash}X
>{\hsize=1.08\hsize\linewidth=\hsize
  \raggedright\arraybackslash}X
>{\hsize=1.09\hsize\linewidth=\hsize
  \raggedright\arraybackslash}X@{}}

\toprule
\textbf{Challenge} &
\textbf{Why It Arises} &
\textbf{Consequence for Active Perception} &
\textbf{Unresolved Gap} \\
\midrule

State estimation in incomplete and changing scenes &
Partial observations, semantic errors, poorly calibrated confidence, and~changing occlusion &
The system may select actions from an incorrect or outdated state and reinforce its initial error &
State representations that preserve task relevance, temporal validity, and~calibrated geometric and semantic~uncertainty \\
\midrule
Action selection under limited time and resources &
Long-term action dependence, large action spaces, and~expensive prediction and feasibility checking &
Greedy decisions can waste the sensing budget, while more extensive planning may become too slow for the task &
Planning methods that adapt their horizon and computational effort to task urgency and available~resources \\
\midrule
Executable sensing on physical platforms &
Ideal viewpoints may ignore reachability, embodiment, motion error, and~interference with the task action &
An informative viewpoint may be unreachable, unsafe, or~detrimental to the subsequent operation &
Joint reasoning about observation value, platform dynamics, task feasibility, and~transfer from simulation to physical~systems \\
\midrule
Collaborative perception under imperfect communication &
Limited bandwidth, transmission delay, pose disagreement, and~agent or sensor failure &
Shared evidence may become outdated or misaligned, while poor coordination causes redundant sensing or observation gaps &
Distributed policies that coordinate motion, sensing, and~message selection while remaining reliable under intermittent~communication \\
\midrule
Safety, privacy, and~system responsibility &
Active movement, physical interaction, camera redirection, zoom, and~persistent observation can affect people and the environment &
The system may cause collision, object damage, privacy intrusion, or~unjustified actions under unresolved uncertainty &
Risk aware stopping, minimum necessary observation, human oversight, and~interpretable reasons for sensing~actions \\
\midrule
Benchmarking, reproducibility, and~fair comparison &
Studies use different scenes, action spaces, sensing budgets, perception backbones, and~model versions &
Reported results are difficult to compare, and~the source of a performance gain or failure often remains unclear &
Shared closed-loop settings, transparent system configurations, and~corresponding simulation and physical~evaluations \\

\bottomrule
\end{tabularx}
\end{adjustwidth}
\end{table}

\subsection{State Estimation in Incomplete and Changing~Scenes}

Active decisions first depend on how the system understands the current scene. The~state must describe not only observed objects and space, but~also which regions remain unknown, which estimates lack reliable evidence, and~whether the remaining uncertainty affects the current task. A~geometric map can represent unobserved space clearly, but~it may not indicate whether that space is relevant to the task. A~semantic map can highlight target objects and task-related regions, but~incorrect labels can direct sensing toward the wrong location. A~compact learned state supports rapid decisions, although~it can make the visual evidence behind an action difficult to~inspect.

The problem becomes more pronounced in open environments. Detectors and semantic models may appear reliable in familiar scenes while becoming overconfident under changing illumination, unfamiliar objects, or~different language descriptions. Once an incorrect prediction is treated as a reliable state, the~system may actively approach an irrelevant region and ignore the target that requires confirmation. Open vocabulary navigation has begun to examine uncertainty caused by prompt and model selection~\cite{bajpai2025uncertainty}, but~the wider connection between semantic confidence and reliable sensing actions remains~unresolved.

Dynamic environments further shorten the period for which a state remains valid. People, vehicles, smoke, deformable objects, and~other robots can move while the sensor is travelling toward the selected view. A~viewpoint that appears useful at the time of planning may lose its value before execution. State estimation must therefore represent not only the current scene, but~also target motion, changing occlusion, and~the possibility that a prediction will become stale. Methods based on static maps and fixed candidate views become unreliable when environmental change approaches or exceeds the speed of action~execution.

{These factors affect the usefulness of a future observation through distinct but interacting mechanisms. Uncertainty in the current state makes the predicted benefit of a candidate view uncertain. Errors in estimated geometry, target position, or~occlusion can change whether the required evidence will be visible. Prediction errors can then cause the system to overestimate how much that observation will reduce uncertainty or support the task. Delays between planning and image acquisition further increase the discrepancy between the predicted and actual observation, because~targets, occluders, and~the observing platform may move during this interval. Changes in the environment or task state can also make an accurately predicted observation less relevant to the decision that follows. Consequently, a~viewpoint that appears informative when selected may provide little useful evidence when reached. Its value should therefore be assessed for the expected acquisition time and updated as new information becomes available, with~replanning when the assumptions supporting the original selection no longer hold.}

\subsection{Action Selection Under Limited Time and~Resources}

The value of a single viewpoint can often be estimated from visibility, coverage, or~expected uncertainty reduction. A~sequence of sensing actions cannot be selected only from the next immediate gain. A~locally informative viewpoint may move the robot away from the task goal or consume the time and energy required for later exploration. Conversely, an~action with limited immediate benefit may expose a new region and place the system in a better position for later decisions. Active reconstruction, target search, and~mobile exploration must therefore balance current observation value with future action opportunities. {This balance should account not only for the expected benefit of an observation, but~also for the time and energy required to reach and acquire it.}

This dependence on future actions rapidly increases computational cost. The~system must compare candidate observations, verify reachability, check safety, and~determine whether the observation can be acquired in time. Continuous positions, camera orientations, imaging parameters, sensor choices, and~communication actions further enlarge the search space. Greedy methods respond quickly, but~they can repeatedly inspect a local region or consume the action budget too early. Longer planning horizons capture future consequences, although~repeated observation prediction and motion checking can prevent real time~execution.

Computational performance also extends beyond neural network inference. Candidate generation, visibility prediction, collision checking, communication delay, and~actuator movement all affect when the new observation becomes available. In~traffic and tracking, an~accurate action can lose its value when it arrives too late. {For example, a~camera orientation selected to reveal a partially occluded target may become ineffective if the target moves behind another obstacle during computation or actuation. Evaluating observation utility should therefore account for the total time required for planning, communication, motion, and~sensing, rather than assuming that the selected view is acquired immediately.} Industrial inspection and offline reconstruction can afford more extensive planning. The~system must also determine when to stop observing. Additional images only increase latency and resource use when they are unlikely to change the grasp, navigation, or~search~decision.

{The stopping condition should therefore be task-dependent rather than defined only by a fixed number of views or an exhausted sensing budget. The~system should continue observing only when the expected marginal benefit of another observation, such as a meaningful reduction in task-relevant uncertainty or a possible change in the downstream decision, justifies the associated time, energy, computation, and~safety costs. Acquisition may stop successfully when the available evidence is sufficient for the task and the selected decision remains stable under plausible alternative observations. It may also stop because candidate views are expected to provide redundant information or because the remaining uncertainty cannot be resolved with the available viewpoints, sensing modalities, or~time. In~the latter case, stopping should not be interpreted as successful perception: the system may instead need to abstain, retreat, execute a conservative action, or~request human assistance. Because~the environment and task state can change, a~stopping decision should also be revisited when new evidence invalidates the assumptions on which it was made.}

\subsection{From Informative Viewpoints to Executable~Actions}

Many methods first select an ideal camera pose in geometric space, but~the physical platform determines whether that pose can be reached. A~manipulator is limited by joint range and self occlusion, a~mobile robot may not fit through the route that provides the desired view, and~a UAV may be unable to remain stable near a structure. A~viewpoint with high predicted value cannot produce useful evidence when it requires unsafe or infeasible~motion.

Simulation often reduces these difficulties. Accurate depth, perfect calibration, and~deterministic motion allow a policy to focus on viewpoint value, but~they can also encourage dependence on information that a physical platform cannot provide. Motion blur, camera vibration, execution delay, and~calibration drift affect both observation quality and action outcome. Randomization and fine tuning with real data can reduce part of this difference, but~they do not replace explicit modelling of platform reachability, sensor placement, and~execution~error.

{Accordingly, the~utility of a future observation should be evaluated jointly with the cost, feasibility, and~risk of obtaining it. A~viewpoint with high expected information or task value may still be undesirable if reaching it requires excessive time or energy, violates the platform's kinematic or dynamic constraints, or~introduces substantial collision risk. Moreover, localization drift, control inaccuracies, actuator uncertainty, and~external disturbances may cause the sensor to reach a pose different from the one assumed during planning, thereby reducing the quality or relevance of the acquired observation. Observation selection should therefore balance expected task benefit against acquisition time, energy consumption, motion feasibility, collision risk, and~the probability and consequences of execution errors.}

Robotic manipulation further shows that sensing actions and task actions cannot always be separated. Moving a wrist camera also changes the manipulator configuration, while rotating an object or removing an occluder changes the task state directly. Recent methods coordinate camera actions and manipulation actions within a shared \mbox{process~\cite{WOS:001454288700001,liu2026sapave,liu2026activevla}.} This integration reduces redundant movement between separate sensing and execution stages, but~it also makes failure more difficult to diagnose. A~failed manipulation may originate from viewpoint selection, state estimation, control error, or~a mismatch among these~components.

{Errors at these stages do not remain isolated but can propagate through the closed perception--action loop. A~sensing or state-estimation error may distort the predicted utility and feasibility of candidate viewpoints, causing the planner to select an uninformative, unsafe, or~difficult-to-reach action. Errors during planning or motion execution may then cause the sensor to reach a pose different from the one assumed during observation prediction, changing the expected visibility, coverage, or~information gain. If~the resulting degraded observation is used to update the scene or target state without accounting for this mismatch, the~estimation error can influence the next round of action selection and may be amplified over successive iterations. Conversely, uncertainty-aware state updating, execution monitoring, and~replanning can interrupt this propagation and allow later observations to correct earlier errors. Robust active perception therefore requires uncertainty and failure information to be communicated across sensing, state estimation, observation prediction, action selection, and~control, rather than treating errors at each stage independently.}

\subsection{Collaborative Perception Under Imperfect~Communication}

Multi agent systems can use sensors at different locations to reduce occlusion and extend the observed area, but~collaboration does not simply require sharing more data. The~agents must determine which member already possesses relevant information, which content is worth transmitting, and~whether the remote observation will remain useful when it arrives. Redundant messages consume bandwidth, while missing a critical observation may cause several agents to enter the same unknown~region.

Where2comm, How2comm, and~ActFormer reduce communication through region or query selection~\cite{hu2022where2comm,yang2023how2comm,huang2024actformer}. These studies show that selective transmission can improve the balance between communication cost and perception quality. Even an accurate message can nevertheless change meaning because of transmission delay and pose error. A~moving target may have left its reported location before the receiver uses the message, while inconsistent localization can prevent observations from different viewpoints from being aligned~correctly.

As the number of agents grows, the~system must jointly determine motion, observation regions, and~communication relationships. Centralized policies can use global information, but~their computational and communication costs increase rapidly. Independent decisions scale more easily, although~they can produce repeated exploration, path conflicts, and~observation gaps. Collaborative active perception must balance local autonomy with global coordination and retain basic task capability when communication becomes intermittent or an agent~fails.

\subsection{Safety, Privacy, and~System~Responsibility}

Active visual perception obtains information through movement and interaction; so, its risks extend beyond incorrect recognition. A~mobile robot may approach a person or obstacle to improve its view, a~UAV may move closer to an unstable structure, and~a manipulator may collide with or damage an object while revealing a hidden state. Observation gain does not by itself justify these actions, particularly when an additional view provides only a limited~improvement.

Applications involving people must also consider whether the observation itself is acceptable. Camera redirection, persistent tracking, and~local zoom can improve pose or identity recognition while collecting information unrelated to the task. A~technically informative viewpoint may invade personal space, and~a high resolution image may contain unnecessary sensitive content. The~system must control not only where it observes, but~also the spatial extent, duration, and~later use of the~observation.

Reliability also depends on whether the system can acknowledge insufficient evidence. An~active policy may continue acquiring views when confidence remains unstable, or~it may force a definite output when reliable evidence is unavailable. A~safe system must distinguish uncertainty that can be resolved through another observation from uncertainty that cannot be resolved under the current conditions. Stopping, retreating, or~requesting human assistance can be more appropriate than continuing when another observation is unlikely to change the decision or requires excessive~risk.

\subsection{Benchmarking, Reproducibility, and~Fair~Comparison}

Existing active visual perception studies use substantially different scenes, action spaces, and~resource constraints. Active reconstruction methods may be evaluated on different object collections and candidate viewpoints, while exploration methods use different map sizes, movement steps, and~stopping conditions. Coverage, success, and~path efficiency can therefore correspond to different levels of task difficulty, which makes direct comparison of final values~misleading.

Fixed images and prerecorded trajectories provide consistent inputs for perception modules, but~they cannot fully compare action selection because different policies cannot change future observations. Interactive simulation supports closed-loop comparison, although~sensor models, action interfaces, and~physical constraints also affect the result. The~lack of corresponding simulation and physical settings makes it difficult to determine whether performance loss originates from the visual model, the~sensing policy, or~motion~control.

Foundation models introduce additional reproducibility problems. Model versions, prompts, external services, and~perception backbones can change over time. Replacing the semantic model may substantially alter the result even when the active decision component remains unchanged. Comparisons in active visual perception must therefore identify the available inputs, action permissions, model versions, and~privileged information rather than recording only the policy name and final~outcome.

These challenges do not occur independently. Incorrect uncertainty changes the predicted value of a viewpoint, computation delay makes the prediction stale, platform constraints prevent execution, and~communication error alters the shared state. The~central challenge is to maintain a reliable perception and action loop when these errors interact, rather than to select a high scoring next observation only under ideal~conditions.

\section{Future~Directions}
\label{sec:future_directions}

Active visual perception is moving from viewpoint control in constrained scenes toward intelligent systems that can adapt their sensing behavior to tasks, environments, and~physical conditions. Future progress will be shaped by advances in task driven learning, multimodal sensing, embodied interaction, collaborative perception, and~responsible deployment. These developments will allow active systems to acquire more relevant visual information, respond more effectively to environmental change, and~operate with greater autonomy in complex real settings. At~the same time, they will require closer coordination among perception, sensing actions, physical execution, and~system~safety.

\subsection{Task Driven Learning and Foundation~Models}

Learning-based methods will continue to play an important role in active visual perception. Reinforcement learning allows a system to improve its sensing strategy through interaction, while self supervised and continual learning can reduce the dependence on manually labelled data. These methods are particularly useful when the relationship between a sensing action and its later task outcome is difficult to describe with a fixed geometric model. A~robot can gradually learn which viewpoints support successful manipulation, while a mobile agent can learn which observations are most useful for navigation and target~search.

Future research will place greater emphasis on observations that are directly related to the current task. A~reconstruction system may not need to observe every surface when only a particular object part is required. A~manipulation system may only need enough information to confirm a grasp region or an insertion point. Target-oriented reconstruction, affordance-guided viewpoint planning, and~active manipulation already show this transition from generic coverage toward task-relevant evidence~\cite{yu2024semantic,hu2026oa,zhang2023affordance,liu2026sapave}.

Foundation models will further extend this ability by connecting visual observations with open vocabulary concepts and language instructions. Systems such as ProFocus, Embodied Reasoner, ActiveVLA, and~SaPaVe indicate that semantic reasoning can help identify which object, region, or~task state requires additional observation~\cite{xue2026profocus,zhang2026embodied,liu2026activevla,liu2026sapave}. However, these models may also produce plausible conclusions that are not supported by the current scene. Future systems will therefore need to distinguish prior knowledge from visual evidence and use new observations to verify uncertain semantic predictions rather than continuing from an unsupported~assumption.

{Here, an~output produced by a foundation model, VLM, or~VLA should be treated as a model-generated hypothesis derived from learned priors and the currently available inputs. It may describe an object, spatial relation, affordance, task state, or~appropriate action even when the relevant scene evidence is occluded, ambiguous, outdated, or~absent. By~contrast, information that is observed and confirmed is grounded in measurements acquired by the sensing system from a task-relevant viewpoint and at the relevant time, and~is supported by checks such as visibility, measurement quality, cross-view or cross-modal consistency, or~agreement with the observed outcome of an executed action. Sensor confirmation does not imply error-free ground truth, but~it provides direct and traceable evidence that a model-generated prediction alone cannot supply. Predictions from foundation models, VLMs, and~VLAs should therefore guide what, where, and~when to observe, while predictions that have not been verified should remain explicitly represented as uncertain hypotheses and should not be incorporated into the confirmed perceptual or task state until sufficient sensory evidence has been obtained.}

\subsection{Multimodal Sensing and Adaptive Sensor~Control}

Future active visual perception systems will rely on a wider range of sensing variables than camera position alone. Modern platforms may contain visual cameras, depth sensors, LiDAR, radar, thermal cameras, and~other sensing devices. Each modality provides different information and performs differently under changes in distance, illumination, weather, and~target motion. Cameras provide detailed appearance, LiDAR supports geometric measurement, radar remains effective under poor visibility, and~thermal sensing can reveal people or objects in~darkness.

{Existing approaches illustrate different levels of spatial integration between learned perception models and sensor data. At~the simpler end, Nowakowski~et~al. converted LiDAR point clouds into flat image-like representations so that pretrained image-based object detection and classification models could be reused, while the laser measurements supplied spatial information such as object distance~\cite{nowakowski2023camera}. This modular strategy can reduce the need to design a new perception model for each sensor representation, although~projecting three-dimensional measurements into a flat view may discard geometric structure and introduce projection-dependent ambiguity. More advanced approaches construct a shared bird's-eye-view representation. For~example, BEVFormer v2 uses perspective supervision and a two-stage architecture in which proposals generated by a perspective-view head support final predictions in the BEV space, enabling modern image backbones to be adapted to multi-camera 3D detection~\cite{yang2023bevformer}. These approaches span a progression from simplified reuse of pretrained image models on projected sensor data to learned spatial representations that support more explicit geometric reasoning. Within~active perception, such fusion or representation mechanisms become observation-selection components when the system adaptively chooses the sensor, viewpoint, sampling configuration, or~representation according to the current task and the evidence that remains uncertain.}

The next stage of multimodal perception will involve more than combining all available sensor data. The~system will need to decide which sensor should be used, when it should operate, and~how its parameters should be adjusted. A~platform may increase image resolution when a small object requires confirmation, raise the frame rate for a rapidly moving target, or~switch to thermal sensing when visible images become unreliable. Multimodal scientific missions already show that observation value depends on both the sensing location and the selected modality~\cite{arora2019multi}.

Joint sensor control can improve robustness and reduce unnecessary computation, but~it also introduces new difficulties. Zooming improves local detail while reducing the field of view. A~higher frame rate provides better temporal information while increasing processing and energy use. Different sensors also require accurate spatial and temporal calibration before their observations can be combined. Future research will need hierarchical sensing policies that first identify the cause of insufficient information and then adjust only the viewpoint, imaging parameter, or~sensing modality that is relevant to the~problem.

\subsection{Embodied Active Perception and~Interaction}

Active visual perception will become more closely integrated with the physical actions of robots. Many current systems still treat sensing as a stage that is completed before manipulation or navigation begins. This separation is difficult to maintain in real environments because every physical action can change the next observation. Moving a wrist camera changes the manipulator configuration, following a visually reliable route changes navigation cost, and~touching or rotating an object changes both its appearance and its physical~state.

Recent active manipulation and vision language action methods have begun to acquire new observations during task execution~\cite{WOS:001454288700001,liu2026activevla,liu2026sapave}. A~robot can adjust its camera when a grasp region becomes occluded, inspect the result of an earlier action, or~interact with an object when camera movement alone cannot reveal the required state. This continuous connection between sensing and acting can reduce unnecessary observation steps and improve adaptation to unexpected~changes.

Physical interaction also introduces greater risk than camera movement. Rotating an object, opening a container, or~removing an occluder may reveal useful information, but~these actions can also disturb the scene or create an irreversible state change. Future systems will need to compare noncontact observation with physical interaction and select the safer option whenever both can provide sufficient evidence. They will also need to consider reachability, collision risk, recovery ability, and~the effect of a sensing action on subsequent task~execution.

\subsection{Collaborative and Distributed~Perception}

Collaborative perception will continue to expand the sensing range of individual robots, vehicles, and~UAVs. Different agents can observe the same environment from complementary locations and share information about targets that remain occluded or distant from one another. In~autonomous driving, vehicles and roadside sensors can provide information beyond the line of sight of the ego vehicle. In~disaster response, multiple UAVs can divide a search region and observe important locations from different~directions.

Existing methods such as Where2comm, How2comm, and~ActFormer have shown that selective communication can reduce data transmission while retaining useful perceptual information~\cite{hu2022where2comm,yang2023how2comm,huang2024actformer}. Future collaborative systems will need to move beyond deciding only what to transmit. They will also need to determine which agent should observe a target, where that agent should move, and~whether the resulting information can still influence the current task~decision.

Communication delay, pose error, and~network failure will remain central obstacles. A~message about a moving object may become outdated before it reaches another agent, while inaccurate relative poses may prevent observations from being aligned correctly. Larger teams also create repeated observations and conflicting movements when coordination is weak. Future distributed policies will need to preserve local autonomy, use communication only when it provides complementary evidence, and~maintain basic task capability when the network becomes~unreliable.

\subsection{Open-World Adaptation and Long-Term~Operation}

Active visual perception systems deployed outside controlled experiments will encounter unfamiliar objects, new layouts, changing weather, sensor degradation, and~different operating conditions. A~sensing policy trained in one environment may continue selecting observations that were useful during training but no longer provide reliable evidence after the environment changes. Long-term operation therefore requires the system to recognize when its current perception and sensing strategy are no longer~suitable.

Active sensing can support adaptation by directing the system toward observations that expose model errors or unfamiliar conditions. A~robot may seek another viewpoint when an object does not match its existing representation, while an environmental monitoring system may revisit a region when the current appearance differs from its historical pattern. Continual learning can then update the perception model or sensing policy from these~observations.

This process must remain controlled. An~unfamiliar state should not automatically trigger aggressive exploration, particularly when physical interaction or movement may create risk. The~system also needs to preserve earlier capabilities rather than repeatedly adapting to recent conditions and forgetting previous environments. Future research will need to connect uncertainty detection, safe exploration, continual learning, and~long-term resource management so that adaptation improves the complete perception and action process rather than only the visual~backbone.

\subsection{Safety, Explainability, and~Responsible~Deployment}

As active visual perception enters autonomous driving, surveillance, healthcare, human robot interaction, and~other sensitive fields, sensing actions will need to satisfy physical, social, and~ethical constraints. An~active system does not only interpret information that has already been collected. It decides where to look, how closely to observe, and~which information should be stored or shared. These decisions can affect personal space, privacy, safety, and~public~trust.

Future systems will need to limit sensing to the information required by the task. A~camera should not continuously zoom toward a person when a lower resolution observation is already sufficient. A~robot should not approach a person or fragile object when the expected information improvement is small. The~system should also be able to stop, retreat, or~request human assistance when uncertainty cannot be resolved~safely.

Explainability will become especially important because an active decision changes which evidence is collected. The~system should be able to indicate which information was missing, why a particular observation was selected, and~whether the acquired evidence changed the decision. These explanations should be linked to the actual perceptual state rather than generated as general language descriptions. Transparent sensing records, privacy-preserving representations, human oversight, and~clear responsibility boundaries will be necessary for dependable~deployment.

Future progress in active visual perception will therefore depend on more than stronger perception models or a larger number of sensing actions. The~field will need systems that understand which evidence matters to the task, select appropriate sensors and viewpoints, coordinate sensing with physical actions, cooperate under imperfect communication, adapt to unfamiliar environments, and~operate within clear safety and ethical~boundaries.

\section{Conclusions}
\label{sec:conclusion}

{This review shows that active visual perception is not simply the processing of multiple observations, but~a closed-loop process that connects task-related information needs, sensing actions, acquired evidence, and~subsequent decisions. Across reconstruction, localization, search, tracking, manipulation, and~multi-agent perception, a~common finding is that an observation is useful only when it provides task-relevant evidence at an acceptable cost and can be acquired reliably under physical and environmental constraints.}

{The reviewed studies further show that uncertainty, delay, execution error, and~environmental change can affect several stages of the perception--action loop, while multi-agent sensing requires coordination of who should observe, what should be communicated, and~whether shared evidence remains timely and spatially aligned. These findings are supported by the cross-task comparisons in Sections~\ref{sec:taxonomy} and \ref{sec:applications}, the~evaluation analysis in Section~\ref{sec:evaluation}, and~the recurring challenges summarized in Section~\ref{sec:challenges}. Future progress therefore requires closer integration of evidence verification, uncertainty-aware planning, executable sensing actions, adaptive stopping, and~reproducible closed-loop evaluation.}

\vspace{6pt} 

\authorcontributions{
Conceptualization, Y.L., S.L. and X.D.;
methodology, Y.L., X.G., H.Z. and J.C.;
investigation, Y.L., X.G., H.Z. and J.C.;
formal analysis, Y.L., X.G., H.Z. and J.C.;
resources, S.L. and X.D.;
writing---original draft preparation, Y.L.;
writing---review and editing, X.G., H.Z., J.C., S.L. and X.D.;
visualization, Y.L., X.G., H.Z. and J.C.;
supervision, S.L. and X.D.;
project administration, S.L. and X.D.;
funding acquisition, S.L. and X.D.
All authors have read and agreed to the published version of the manuscript.
}

\funding{\hl{This} 
 research was funded by the Guangxi Science and Technology Program (Nos. Guike GXYD2504850002 and Guike XT2600870042).
}

\institutionalreview{Not applicable.}

\informedconsent{Not applicable.}

\dataavailability{No new data were created or analyzed in this study. Data sharing is not applicable to this article.}

\acknowledgments{During the preparation of this manuscript, the~authors used \hl{ChatGPT} 
 (GPT-5, OpenAI, San Francisco, CA, USA) solely for English-language polishing and grammatical improvement. All AI-assisted language edits were carefully reviewed and revised by the authors, who take full responsibility for the content of this~manuscript.}

\conflictsofinterest{The authors declare no conflicts of~interest.}

\begin{adjustwidth}{-\extralength}{0cm}

\reftitle{References}


\PublishersNote{}
\end{adjustwidth}
\end{document}